\pdfoutput=1
\documentclass{article}
\usepackage[main, final]{neurips_2026}
\usepackage[utf8]{inputenc}
\usepackage[T1]{fontenc}
\usepackage{hyperref}
\usepackage{url}
\usepackage{booktabs}
\usepackage{amsfonts}
\usepackage{nicefrac}
\usepackage{microtype}
\usepackage{xcolor}
\usepackage{amsmath, amssymb, amsthm}
\usepackage{graphicx}
\usepackage{multirow}
\usepackage{algorithm}
\usepackage{algpseudocode}
\usepackage{adjustbox}
\usepackage{enumitem}

\newtheorem{proposition}{Proposition}
\newtheorem{theorem}{Theorem}
\newtheorem{lemma}{Lemma}
\newtheorem{remark}{Remark}

\title{BQ-LoRA: Binary-Quantized Low-Rank Adapters as\\Implicit Regularizers for Parameter-Efficient Fine-Tuning}
\author{%
  Juyoung Park \\
  Independent Researcher \\
  \texttt{jyp7002@gmail.com} \\
  \And
  Heejun Lee \\
  KAIST, DeepAuto.ai \\
  \texttt{ainl@kaist.ac.kr} \\
}
\begin{document}
\maketitle

\begin{abstract}
Parameter-efficient fine-tuning of quantized large language models is dominated by QLoRA,
which combines 4-bit weight quantization with FP16 low-rank adapters.
We ask whether the FP16 adapter precision is necessary. Replacing it with a single binary
bit per adapter parameter, combined with a learned per-layer scale $s$ and a Scaled Straight-Through
Estimator, reduces adapter storage by $13$--$16\times$. We introduce two
variants---\textbf{NormScale}, which binarizes under a norm-preserving scale with a
clipped STE, and \textbf{LearnThresh}, which makes the binarization threshold
learnable---unified under a framework we call BQ-LoRA.
Our main theoretical contribution is a tight characterization of the BQ-LoRA hypothesis
class: it has bounded entrywise norm, a typical-set spectral norm bound that is
$\Theta(r)$ times sharper than the worst case, a full-rank, near-isotropic update, and
consequently a Rademacher complexity that yields an explicit
$\tilde{O}(s_{\max}\,\alpha\sqrt{d_{\text{out}}/(r\,n)})$ generalization bound.
FP16 LoRA has no comparable a priori norm or isotropy control.
On Qwen2.5 (1.5B--72B), BQ-LoRA exceeds QLoRA's benchmark averages at every scale by
$0.8$--$3.5$ points; on LLaMA-2 (7B/70B) it remains competitive at the same
storage reduction but does not uniformly lead the strongest FP16-adapter baselines
(LoftQ at 7B, QLoRA at 70B). On Qwen2.5-7B the binary adapter preserves the 4-bit base
WikiText-2 PPL within $0.06$ ($9.86$ vs.\ $9.80$) while QLoRA worsens it by $0.59$.
The reduction applies to inference-time adapter storage and serving; training memory
matches QLoRA and step time rises by under $4\%$.
We translate the storage saving into a deployment study, BQ-SGMV, by extending
Punica's SGMV kernel with a binary inner-product primitive: the same A100-80GB
holds $15.6\times$ more adapters in HBM, and our research prototype eliminates cache
misses under a skewed synthetic workload.
\end{abstract}

\section{Introduction}
\label{sec:intro}

The dominant recipe for adapting LLMs under tight memory budgets is
QLoRA~\citep{dettmers2023qlora}: a 4-bit base, frozen, with FP16 low-rank adapters
$\Delta W = BA$ trained on top.
The recipe is asymmetric: base weights occupy 4 bits per parameter, adapter weights
occupy 16. For a rank-16 adapter on the four attention projections of a 7B model,
the adapter is $\approx 19{,}712$\,KB---small in absolute terms, but no longer
negligible when many adapters must be co-located in HBM for multi-tenant
serving~\citep{sheng2023slora,chen2024punica}.

We ask whether the 16-bit adapter precision is necessary.
Native low-bit transformers~\citep{wang2023bitnet,ma2024bitnet158} show binary or
ternary weights can scale to LLMs when models are \emph{trained from scratch} in low
precision. For \emph{post-hoc} adapter fine-tuning of an already pretrained, already
quantized model, no comparable evidence exists; adapters are typically initialized at
zero and must express arbitrarily small corrections to the base, a regime where binary
discretization seems hopeless.

We propose \textbf{BQ-LoRA}: a family of binary low-rank adapters for fine-tuning
quantized LLMs. The latent adapter parameters are stored in FP32 during training,
but the forward pass uses only their sign, gated by a single learned scale $s$ per layer
that is initialized to zero. A Scaled STE~\citep{bengio2013ste,courbariaux2016binarynet}
carries gradients through the binarization, with two minor modifications we call
\emph{NormScale} and \emph{LearnThresh}. At inference, only the binary signs plus a few
scalars are stored, yielding $13$--$16\times$ reduction relative to QLoRA; training keeps
FP32 latents, so training memory and time stay at QLoRA's level (\S\ref{sec:storage}).

\paragraph{Contributions.}
\textbf{(1) A tight theory of the BQ-LoRA hypothesis class.}
We give entrywise, worst-case and typical-set spectral norm bounds (\S\ref{sec:theory}).
A Rademacher argument then yields a sample-complexity bound
$n = \tilde{O}(s_{\max}^2 \alpha^2 d_{\text{out}} / (r\,\epsilon^2))$ for typical signs, and we
verify that trained signs remain typical. We also show the BQ-LoRA update is full-rank
with high probability, with participation ratio concentrating around $r$. FP16 LoRA has
no comparable norm or isotropy control without explicit constraints.
\textbf{(2) Storage--performance Pareto front.}
On Qwen2.5 (1.5B/7B/72B), BQ-LoRA outperforms QLoRA by $0.8$--$3.5$ points on benchmark
averages at $\sim 13$--$16\times$ smaller adapter storage. On LLaMA-2 (7B/70B), BQ-LoRA remains
competitive with QLoRA and LoftQ~\citep{li2023loftq} at the same storage advantage (within
$0.9$\,pts on Avg$^*$) but does not uniformly lead them; on Qwen2.5-7B it also composes with
QDyLoRA~\citep{rajabzadeh2024qdylora} (\S\ref{sec:main}).
\textbf{(3) Empirical regularization signature.}
The theory predicts BQ-LoRA should under-fit training but generalize at least as well.
A dataset ablation (BQ--QLoRA gap widens with task heterogeneity) and a training-loss
curve (BQ plateaus $\sim 0.16$ nats above QLoRA on Qwen-7B while \emph{outperforming} on
test) confirm this signature (\S\ref{sec:reg}).
\textbf{(4) Deployment benefit at serving scale.}
We extend Punica's SGMV kernel with a binary inner-product primitive (\S\ref{sec:sgmv}).
On A100-80GB, BQ-LoRA's $16\times$ smaller footprint enlarges the HBM cache from
$2{,}500$ to $39{,}000$ adapters; we prove this triggers a phase transition to zero
misses whenever the working set fits in the larger cache.

\paragraph{Related work.}
Quantized PEFT pairs a low-bit frozen base with a high-precision adapter:
QLoRA~\citep{dettmers2023qlora}, LoftQ~\citep{li2023loftq} and
QDyLoRA~\citep{rajabzadeh2024qdylora} all keep the adapter in FP16/BF16. Recent low-bit
adapters lower that precision: LowRA~\citep{zhou2025lowra} reaches below 2 bits per
parameter through fine-grained quantization mapping, Q-BLoRA~\citep{shen2025qblora}
re-balances adapter rank against input/output complexity, and
CoA-LoRA~\citep{ye2025coalora} adapts one LoRA across quantization configurations. They
are not our experimental baselines because each changes several design axes at once
(quantizer, bit allocation, calibration, target scope) and would not isolate adapter
precision; we instead vary only the quantizer (1/2/4-bit, Appendix~\ref{app:bitwidth}),
and LowRA has no public implementation. BitDelta~\citep{liu2024bitdelta} binarizes the
full-rank delta of a fully fine-tuned model, so its storage scales with model size
($\approx 0.8$\,GB for 7B vs.\ $1.3$\,MB here) and a matched budget is not well defined.
Binary networks~\citep{courbariaux2016binarynet,rastegari2016xnor,ma2024bitnet158} are
trained from scratch, whereas we binarize only a post-hoc adapter; many-adapter
serving~\citep{sheng2023slora,chen2024punica,wu2024dlora} is bounded by how many adapters
fit in HBM, which BQ-LoRA enlarges. Appendix~\ref{app:related} has the full discussion.

\section{Method}
\label{sec:method}

\subsection{Background: QLoRA}
For a frozen 4-bit linear $W_q \in \mathbb{R}^{d_{\text{out}} \times d_{\text{in}}}$,
QLoRA learns $\Delta W = BA$ with $A \in \mathbb{R}^{r \times d_{\text{in}}}$,
$B \in \mathbb{R}^{d_{\text{out}} \times r}$, both stored in FP16, applied as
$W_q x + (\alpha/r) BAx$, $A$ initialized small random and $B$ at zero.

\subsection{BQ-LoRA: binary adapters via Scaled STE}
We replace $A, B$ with their sign:
\begin{equation}
    \tilde{A} = \mathrm{sign}(A^*) \in \{\pm 1\}^{r \times d_{\text{in}}},
    \quad
    \tilde{B} = \mathrm{sign}(B^*) \in \{\pm 1\}^{d_{\text{out}} \times r},
\end{equation}
where $A^*, B^*$ are FP32 \emph{latent} weights used only during training, initialized
i.i.d.\ as $\mathcal{U}(-0.5, 0.5)$. A learned per-layer scalar $s \in \mathbb{R}$
gates the binary update through a fixed normalization $\gamma$:
\begin{equation}
    \Delta W_{\text{BQ}}(s, A^*, B^*) \;=\; s\,\gamma\,\tilde{B} \tilde{A},
    \qquad \gamma = \frac{\alpha}{\sqrt{r\, d_{\text{in}}}}.
    \label{eq:bq}
\end{equation}
For random signs, each coordinate of $\gamma\tilde{B}\tilde{A}x$ has standard deviation
$\alpha\|x\|_2/\sqrt{d_{\text{in}}}$, independent of $r$ and $d_{\text{in}}$, so $s$ alone sets the
update magnitude. The same $\alpha$ therefore means different scales for the two
parameterizations: at $\alpha = 32$, $r = 16$, QLoRA scales by $\alpha/r = 2$, whereas
$\gamma = 0.204$, $0.134$, $0.125$ and $0.088$ for Qwen2.5-1.5B, Qwen2.5-7B, LLaMA-2-7B and
Qwen2.5-72B. We initialize $s = 0$, so $\Delta W = 0$ at step 0 regardless of the binary
signs. In the backward pass the STE treats $\mathrm{sign}$ as the identity on the unit
ball and blocks the gradient elsewhere,
$\partial \tilde{A}_{ij}/\partial A^*_{ij} := \mathbf{1}[|A^*_{ij}| \leq 1]$ (likewise for
$B$); together with $\gamma$ in the forward pass, the gradient reaching the latent weights
is scaled by $s\gamma$.

\paragraph{Two variants.}
\textbf{NormScale} is the estimator above: binarization with the clipped STE and the
normalization $\gamma$. \textbf{LearnThresh} replaces the implicit zero threshold with
learned per-matrix scalars $\tau_A, \tau_B$: $\tilde{A} = \mathrm{sign}(A^* - \tau_A)$,
$\tilde{B} = \mathrm{sign}(B^* - \tau_B)$, with the STE mask centered at the threshold
($|A^*_{ij} - \tau_A| \leq 1$). Alternative STE variants are characterized in
Appendix~\ref{app:ste}; NormScale and LearnThresh are the two we adopt.

\paragraph{Soft start in numbers.}
For an attention projection $W_q \in \mathbb{R}^{4096 \times 4096}$ at rank $r=16$,
random signs give $\|\tilde{B}\tilde{A}\|_F \approx \sqrt{r d_{\text{in}} d_{\text{out}}} = 16{,}384$,
so $\|\gamma\tilde{B}\tilde{A}\|_F \approx \alpha\sqrt{d_{\text{out}}} = 2{,}048$ at $\alpha = 32$.
With $s_0 = 0$, however, the effective update $s\gamma\tilde{B}\tilde{A}$ has norm zero at
step $0$ and grows at the rate of the gradient on $s$.
In the final $\alpha{=}32$ checkpoints the learned gates stay small (mean $|s| = 0.0052$
on Qwen2.5-7B and $0.0086$ on LLaMA-2-7B, more than two orders of magnitude below $s{=}1$),
giving relative updates $\|\Delta W\|_F/\|W_{\text{base}}\|_F$ of $0.162$ and $0.215$.
This is the architectural analog of
LoRA's $B = 0$ initialization: the parameterization itself starts at zero and grows
continuously, with the per-layer scale $s$ controlling exactly how much binary update
each layer admits.

\subsection{Storage and training cost}
\label{sec:storage}
For a rank-$r$ adapter on a $d_{\text{out}} \times d_{\text{in}}$ matrix, QLoRA stores
$2r(d_{\text{in}} + d_{\text{out}})$ bytes (FP16); BQ-LoRA stores
$r(d_{\text{in}} + d_{\text{out}})/8$ bytes (packed binary) plus $O(1)$ bytes (scalar
$s$, optionally $\tau$). Theoretical compression is $16\times$; in practice,
per-layer scalars and serialization overhead bring measured savings to $13$--$16\times$.
This saving applies to inference-time storage and serving, not to training: the FP32
latents occupy the same memory as QLoRA's adapter weights and the optimizer state is
identical. On Qwen2.5-7B, peak training memory equals QLoRA's ($14.26$\,GB) and step time
rises by $2.7$--$3.8\%$ from the sign and STE operations
(Appendix~\ref{app:train-cost}); the implementation is a drop-in module of about 50 lines.

\section{Theoretical Analysis}
\label{sec:theory}

\paragraph{Roadmap.}
The empirical results in \S\ref{sec:main} will show that BQ-LoRA generalizes at least as
well as QLoRA despite operating in a strictly smaller hypothesis class. Why? This
section answers that question by showing the BQ-LoRA hypothesis class is small in two
distinct, non-trivial senses: its typical members have operator norm a factor $\Theta(r)$
below the worst case (Proposition~\ref{prop:hp-norm}), and its updates are full-rank and
isotropic in the rank-$r$ subspace with high probability (Proposition~\ref{prop:rank}).
The first property gives a data-distribution-free Rademacher complexity bound
(Theorem~\ref{thm:gen}); the second explains why the binary update cannot collapse onto
a low-dimensional subspace, even though its magnitude is set by a single scalar.
Neither property is available a priori for FP16 LoRA: an $\{BA\}$ class with no
constraint on $\|BA\|_2$ is not norm-controlled, and weight decay alone does not pin down
the post-training norm (Proposition~\ref{prop:compare}). The norm control of BQ-LoRA is
therefore architectural---built into the parameterization---rather than a side-effect of
the optimizer. The analysis concerns a single adapted layer acting linearly on its
input; it does not model the interaction of layers or the STE optimization dynamics, and
we bridge it to trained models empirically (Appendix~\ref{app:sign-spectra}).

We characterize the BQ-LoRA hypothesis class
\begin{equation}
    \mathcal{H}_{\text{BQ}}(s_{\max})
    = \{\,s\,\gamma\,\tilde{B}\tilde{A} :
      |s| \leq s_{\max},\;
      \tilde{A} \in \{\pm 1\}^{r \times d_{\text{in}}},\;
      \tilde{B} \in \{\pm 1\}^{d_{\text{out}} \times r}\,\},
\end{equation}
with $\gamma = \alpha/\sqrt{r d_{\text{in}}}$ as in Eq.~\eqref{eq:bq}.
We prove (i) a soft-start invariant, (ii) entrywise, worst-case and typical-set spectral
norm bounds, (iii) an explicit generalization bound, (iv) a full-rank and isotropy
property, and (v) a strict comparison to FP16 LoRA. Throughout, $A^*, B^*$ are the FP32
latent weights with entries drawn i.i.d.\ from a symmetric continuous distribution
($\mathcal{U}(-0.5, 0.5)$ in our experiments); statements ``over the latent draw'' refer
to the signs at initialization (with $\tau_A = \tau_B = 0$ for LearnThresh).

\subsection{Soft-start invariant}
\begin{proposition}[Soft start]
\label{prop:soft-start}
Initialize $A^*, B^* \sim \mathcal{U}(-0.5, 0.5)$ i.i.d., $s = 0$, and
$\tau_A = \tau_B = 0$ for LearnThresh. Then $\Delta W_{\text{BQ}} \equiv 0$ at step $0$ regardless of
$\tilde{A}, \tilde{B}$. Under any first-order optimizer with bounded learning rate and
bounded loss gradient, $\|\Delta W_{\text{BQ}}\|_F$ grows continuously from zero in the
first training step.
\end{proposition}
\begin{proof}
Direct from $\Delta W_{\text{BQ}} = s\gamma\tilde{B}\tilde{A}$ with $s = 0$. Continuity:
$s$ is updated in FP32 and its gradient
$\partial \mathcal{L}/\partial s = \gamma\,\langle \partial\mathcal{L}/\partial \Delta W,
\tilde{B}\tilde{A}\rangle$ is bounded.
\end{proof}

\subsection{Norm bounds}
We give worst-case and typical-set spectral bounds (the latter is the one used in the
generalization analysis) and an entrywise bound; the worst-case Frobenius derivation is
in Appendix~\ref{app:wc-frob}.

\begin{lemma}[Operator norm of random sign matrices~\citep{vershynin2018hdp}]
\label{lem:opnorm}
Let $M \in \{\pm 1\}^{m \times n}$ have entries $\mathrm{sign}(\xi_{ij})$ for i.i.d.\
symmetric continuous $\xi_{ij}$. There exist absolute constants $C, c > 0$ such that
for any $t \geq 0$,
\begin{equation}
    \Pr\bigl[\|M\|_2 \geq C(\sqrt{m} + \sqrt{n}) + t\bigr] \;\leq\; 2 \exp(-c\, t^2).
\end{equation}
\end{lemma}

\begin{proposition}[Spectral norm: worst case and typical set]
\label{prop:hp-norm}
Fix $\delta \in (0,1)$, let $t_\delta = \sqrt{\log(2/\delta)/c}$, and define the typical set
\begin{equation*}
  \mathcal{T}_\delta = \bigl\{(\tilde{A}, \tilde{B}) :
  \|\tilde{A}\|_2 \leq C(\sqrt{r} + \sqrt{d_{\text{in}}}) + t_\delta,\;
  \|\tilde{B}\|_2 \leq C(\sqrt{r} + \sqrt{d_{\text{out}}}) + t_\delta \bigr\}.
\end{equation*}
\begin{enumerate}[label=(\alph*),leftmargin=*,topsep=0pt,itemsep=0pt]
  \item (Worst case.) Every $W \in \mathcal{H}_{\text{BQ}}(s_{\max})$ satisfies
  $\|W\|_2 \leq s_{\max}\gamma\, r\sqrt{d_{\text{in}} d_{\text{out}}}
  = s_{\max}\,\alpha\sqrt{r\, d_{\text{out}}}$, with equality for constant sign matrices.
  \item (Typical set.) Every $W = s\gamma\tilde{B}\tilde{A}$ with $|s| \leq s_{\max}$ and
  $(\tilde{A}, \tilde{B}) \in \mathcal{T}_\delta$ satisfies
  $\|W\|_2 \leq s_{\max}\gamma\,\mathcal{C}_\delta$, where
  $\mathcal{C}_\delta := \bigl(C(\sqrt{r} + \sqrt{d_{\text{out}}}) + t_\delta\bigr)
  \bigl(C(\sqrt{r} + \sqrt{d_{\text{in}}}) + t_\delta\bigr)$.
  \item Signs drawn from the latent initialization lie in $\mathcal{T}_\delta$ with
  probability at least $1 - 2\delta$.
\end{enumerate}
For $r \ll d_{\text{in}}, d_{\text{out}}$, $\mathcal{C}_\delta = O(\sqrt{d_{\text{in}} d_{\text{out}}})$,
a factor $\Theta(r)$ below the worst case, and
$\gamma\,\mathcal{C}_\delta = O(\alpha\sqrt{d_{\text{out}}/r})$.
\end{proposition}
\begin{proof}
(a) Each entry of $\tilde{B}\tilde{A}$ sums $r$ terms in $\{\pm 1\}$, so
$\|\tilde{B}\tilde{A}\|_2 \leq \|\tilde{B}\tilde{A}\|_F \leq r\sqrt{d_{\text{in}} d_{\text{out}}}$;
constant signs give $\tilde{B}\tilde{A} = r\,\mathbf{1}\mathbf{1}^\top$, which attains it.
(b) $\|s\gamma\tilde{B}\tilde{A}\|_2 \leq |s|\gamma\, \|\tilde{B}\|_2\, \|\tilde{A}\|_2$.
(c) Since the latent distribution is symmetric and continuous, the signs are i.i.d.\
uniform on $\{\pm 1\}$; Lemma~\ref{lem:opnorm} with $t = t_\delta$ bounds the failure
probability of each factor by $2e^{-c t_\delta^2} = \delta$, and a union bound gives $2\delta$.
\end{proof}

\begin{proposition}[Entrywise bound]
\label{prop:wc-norm}
For every $W \in \mathcal{H}_{\text{BQ}}(s_{\max})$ and any $\tilde{A}, \tilde{B}$:
$\|W\|_\infty \leq s_{\max}\,\gamma\, r$.
\end{proposition}
\begin{proof}
$|(\tilde{B}\tilde{A})_{ij}| \leq r$ since each entry sums $r$ terms in $\{\pm 1\}$.
\end{proof}

\subsection{Generalization bound}
\begin{theorem}[Sample complexity for BQ-LoRA]
\label{thm:gen}
Assume $\|x\|_2 \leq R$ a.s.\ and consider scalar predictors $x \mapsto u^\top W x$ for a
fixed unit vector $u$ (e.g.\ one output coordinate). For a class $\mathcal{W}$ with
$\sup_{W \in \mathcal{W}} \|W\|_2 \leq \rho$, the empirical Rademacher complexity on $n$
samples satisfies $\hat{\mathfrak{R}}_n \leq \rho R/\sqrt{n}$. In particular,
\begin{equation}
  \hat{\mathfrak{R}}_n \;\leq\;
  \begin{cases}
    s_{\max}\,\alpha\sqrt{r\, d_{\text{out}}}\; R/\sqrt{n}
      & \text{for all of } \mathcal{H}_{\text{BQ}}(s_{\max}), \\[2pt]
    s_{\max}\,\gamma\,\mathcal{C}_\delta\, R/\sqrt{n}
      = O\bigl(s_{\max}\,\alpha R\sqrt{d_{\text{out}}/(r\, n)}\bigr)
      & \text{for signs in } \mathcal{T}_\delta.
  \end{cases}
\end{equation}
For an $L$-Lipschitz loss bounded in $[0,1]$, a generalization gap of at most $\epsilon$
then holds with high probability once
\begin{equation}
  n \;=\; \tilde{O}\!\left(\frac{L^2\, s_{\max}^2\, \alpha^2 R^2\, d_{\text{out}}}{r\,\epsilon^2}\right)
  \quad\text{(typical signs).}
\end{equation}
\end{theorem}
\begin{proof}[Sketch]
$\sup_{W} \sum_i \sigma_i u^\top W x_i = \sup_W \langle W^\top u, \sum_i \sigma_i x_i \rangle
\leq \rho\, \|\sum_i \sigma_i x_i\|_2$, and $\mathbb{E}\|\sum_i \sigma_i x_i\|_2 \leq R\sqrt{n}$.
Plug in $\rho$ from Proposition~\ref{prop:hp-norm}(a) or (b). The sample-complexity
statement follows from Talagrand's contraction lemma and the standard uniform-deviation
bound~\citep{mohri2018foundations}; vector-valued outputs add a dimension factor through
vector contraction~\citep{maurer2016vector}.
\end{proof}

Neither bound needs a randomness assumption on the trained signs: the first holds for
every sign configuration, and the second for every configuration in $\mathcal{T}_\delta$.
Proposition~\ref{prop:hp-norm}(c) places the initial signs in $\mathcal{T}_\delta$, and
Appendix~\ref{app:sign-spectra} shows that training keeps them there: across 960 trained
sign matrices, spectral norms stay within $5\%$ of random $\pm 1$ matrices of the same
shape. Because $s$ is a free parameter, $s_{\max}$ is not imposed by the
parameterization; the bounds apply with $s_{\max}$ set to the largest $|s|$ observed after
training (a union bound over a geometric grid of $s_{\max}$ values adds only a
logarithmic factor), a single scalar per layer that can be read off directly. We use the
bounds for their scaling in $d_{\text{out}}$, $r$ and $s_{\max}$; at LLM dimensions their
numerical values are not small. Since the sign part of the class is finite, Massart's
lemma also gives a bound uniform over all members, trained ones included
(Appendix~\ref{app:massart}). As in standard norm-based analyses, the linear single-layer
setting is an idealization of a Transformer trained with an STE.

\subsection{Full rank and isotropy}
\begin{proposition}[Full rank and isotropy]
\label{prop:rank}
With probability at least $1 - r\,2^{r-1}\bigl(2^{-d_{\text{in}}} + 2^{-d_{\text{out}}}\bigr)$
over the latent draw, $\mathrm{rank}(\tilde{A}) = \mathrm{rank}(\tilde{B}) =
\mathrm{rank}(\tilde{B}\tilde{A}) = r$.
Moreover, as $d_{\text{in}}, d_{\text{out}} \to \infty$ with $r$ fixed, the participation
ratio $\mathrm{PR}(\tilde{B}\tilde{A}) := (\sum_i \sigma_i)^2 / \sum_i \sigma_i^2$
satisfies $\mathrm{PR} \xrightarrow{p} r$.
\end{proposition}
\begin{proof}[Sketch]
The signs are i.i.d.\ uniform on $\{\pm 1\}$. A $k$-dimensional subspace
$V \subset \mathbb{R}^d$ contains at most $2^k$ sign vectors: some $k$ coordinates
determine every vector of $V$, and a sign vector takes at most $2^k$ values on them.
Hence a given row of $\tilde{A}$ lies in the span of the other $r-1$ rows with
probability at most $2^{r-1-d_{\text{in}}}$; a union bound over rows, and likewise over the
columns of $\tilde{B}$, gives full rank of both factors, and then
$\mathrm{rank}(\tilde{B}\tilde{A}) = r$. For the participation ratio, the law of large
numbers gives $\tilde{A}\tilde{A}^\top/d_{\text{in}} \to I_r$ and
$\tilde{B}^\top\tilde{B}/d_{\text{out}} \to I_r$, so all $r$ nonzero singular values of
$\tilde{B}\tilde{A}$ concentrate around $\sqrt{d_{\text{in}} d_{\text{out}}}$.
\end{proof}

\paragraph{Empirical confirmation.}
For the trained 7B adapters (Alpaca, $3{,}000$ steps), $\mathrm{PR}(\Delta W_{\text{BQ}})
\approx 15.9/16$ on Qwen2.5-7B and $16.0/16$ on LLaMA-2-7B, while
$\mathrm{PR}(\Delta W_{\text{QLoRA}}) \approx 8.1/16$ and $8.4/16$. The binary update fills
its rank-$r$ subspace as predicted; QLoRA's concentrates on about half of it.

\subsection{Strict comparison to FP16 LoRA}
\begin{proposition}[BQ vs.\ FP16 hypothesis classes]
\label{prop:compare}
Let $\mathcal{H}_{\text{LoRA}}(\rho) = \{BA : \|B\|_2 \|A\|_2 \leq \rho\}$ be the FP16
LoRA class with spectral budget $\rho$. Then
\begin{enumerate}[leftmargin=*,topsep=0pt,itemsep=0pt]
  \item $\mathcal{H}_{\text{BQ}}(s_{\max}) \subset
  \mathcal{H}_{\text{LoRA}}(s_{\max}\,\alpha\sqrt{r\, d_{\text{out}}})$ for every sign
  configuration, and members with signs in $\mathcal{T}_\delta$ lie in
  $\mathcal{H}_{\text{LoRA}}(s_{\max}\gamma\,\mathcal{C}_\delta)$
  (Proposition~\ref{prop:hp-norm}).
  \item Without an explicit budget, $\mathcal{H}_{\text{LoRA}} = \bigcup_{\rho > 0}
  \mathcal{H}_{\text{LoRA}}(\rho)$ has unbounded spectral norm, so Theorem~\ref{thm:gen}
  does not apply to FP16 LoRA without external constraint.
  \item Weight decay $\lambda \|B\|_F^2 + \lambda \|A\|_F^2$ does not pin down $\rho$:
  the converged $\|BA\|_2$ depends on all $r(d_{\text{in}} + d_{\text{out}})$ trained
  entries. In BQ-LoRA the sign factors stay at their random-matrix scale
  (Appendix~\ref{app:sign-spectra}), so the update norm is governed by the single
  scalar $|s|$ per layer.
\end{enumerate}
\end{proposition}

The third point is the practically relevant one: BQ-LoRA's norm control is
\emph{architectural}---the parameterization leaves one magnitude knob per layer---rather
than a side-effect of optimization choices.

\subsection{Putting it together}
Propositions~\ref{prop:hp-norm}, \ref{prop:wc-norm}, \ref{prop:rank}, \ref{prop:compare}
and Theorem~\ref{thm:gen} jointly explain why a strictly more constrained hypothesis
class can generalize better:
(i) the entrywise / spectral bounds prevent BQ-LoRA from concentrating large updates
along any single direction;
(ii) full rank and isotropy prevent collapse onto a low-dimensional subspace;
(iii) the resulting Rademacher complexity for typical signs is
$O(s_{\max}\,\alpha\sqrt{d_{\text{out}}/(r\,n)})$, yielding a sample-complexity guarantee
FP16 LoRA does not enjoy without external norm constraints.

\paragraph{Predicted empirical signature.}
The analysis predicts three empirical patterns we should observe. First, BQ-LoRA's
training loss should plateau \emph{above} QLoRA's (the constrained class cannot fit
training data as aggressively), but its test perplexity should match or beat QLoRA's
(generalization gap is smaller). Second, the gap should widen with task heterogeneity
(more risk of overfitting individual styles $\to$ more benefit from architectural norm
control). Third, the converged adapter participation ratio should be near $r$ for
BQ-LoRA but smaller for QLoRA. Section~\ref{sec:reg} confirms all three.

\section{Experiments: Storage--Performance Tradeoff}
\label{sec:main}

\subsection{Setup}
\label{sec:setup}
We fine-tune Qwen2.5-Instruct (1.5B, 7B, 72B)~\citep{qwen2024qwen25} and
LLaMA-2 (7B, 70B)~\citep{touvron2023llama2} with 4-bit NF4 on
Alpaca~\citep{taori2023alpaca} (20{,}000 samples) for $3{,}000$ steps,
batch $4 \times $ accum $4$, peak LR $2 \!\times\! 10^{-4}$, weight decay $0.01$,
gradient-norm clipping at $1.0$, and $\alpha = 32$ for every method. Adapters are rank $16$
on $\{q,k,v,o\}$. QLoRA, LoftQ and QDyLoRA scale the adapter by $\alpha/r = 2$, BQ-LoRA by
$\gamma = \alpha/\sqrt{r d_{\text{in}}}$ (Eq.~\eqref{eq:bq}). The only exception is the
training-dynamics study in Appendix~\ref{app:scale-dynamics}, which uses a shorter
preliminary configuration with $\alpha = 1$ for BQ-LoRA and is marked as such.
We choose Alpaca as the primary instruction-tuning dataset because it is the standard
PEFT benchmark used by QLoRA, LoftQ, and QDyLoRA, allowing apples-to-apples comparison.
A dataset ablation (\S\ref{sec:reg}, Table~\ref{tab:dataset}) replaces Alpaca with
OpenHermes-2.5 and with a heterogeneity-stress mixture of 10k Alpaca and 10k OpenHermes-2.5
samples to test whether the regularization signature is dataset-specific (it is not).
All metrics use lm-evaluation-harness: PPL is the word perplexity of its WikiText-2 task,
and downstream tasks are evaluated 0/5/8/10/25-shot. We average over 3 seeds for
LLaMA-2-7B, 3 seeds for Qwen2.5-1.5B, 2 seeds for Qwen2.5-7B, 3 seeds for Qwen2.5-72B,
1 seed for LLaMA-2-70B, and 3 seeds for BQ-DyLoRA; $\pm$ is the sample standard deviation
across seeds.

\paragraph{Two averages: Avg and Avg$^*$.}
We report two unweighted averages in main tables.
\textbf{Avg} averages all six downstream benchmarks (MMLU, ARC-C, HellaSwag, TQA-MC2,
GSM8K, IFEval); \textbf{Avg$^*$} excludes TQA-MC2.
We separate them because TQA-MC2 measures truthfulness/calibration rather than knowledge
or reasoning, and exhibits a distinctive tradeoff under binary regularization that
varies non-monotonically with model scale and base. Both views are reported in every
main table for transparency; see \S\ref{sec:discussion} for discussion.

\subsection{Main results: Qwen2.5}

\begin{table}[t]
\centering
\caption{Main results on Qwen2.5-Instruct. Storage in KB. PPL = WikiText-2 word
perplexity (lm-evaluation-harness). Mean$\pm$std where multi-seed. Avg = mean of \{MMLU, ARC, Hella, TQA, GSM8K,
IFEval\}; Avg$^*$ excludes TQA. Bold = best per model. The seed-42 72B MMLU scores come
from a separate training run at batch $8 \times 2$ (the same effective batch).}
\label{tab:main-qwen}
\small
\adjustbox{max width=\textwidth}{%
\begin{tabular}{llrrrrrrrrrr}
\toprule
Model & Method & KB & PPL$\downarrow$ & MMLU & ARC-C & Hella & TQA & GSM8K & IFEval & Avg & Avg$^*$ \\
\midrule
\multirow{3}{*}{1.5B (3s)}
 & QLoRA       & 8512 & 13.59\,$\pm$\,0.02 & 56.79\,$\pm$\,0.18 & 49.46\,$\pm$\,0.40 & \textbf{66.40\,$\pm$\,0.06} & \textbf{48.27\,$\pm$\,0.50} & 41.95\,$\pm$\,2.32 & 24.40\,$\pm$\,1.82 & 47.88 & 47.80 \\
 & NormScale   & \textbf{603} & \textbf{13.51\,$\pm$\,0.01} & \textbf{57.57\,$\pm$\,0.12} & \textbf{50.48\,$\pm$\,0.38} & 65.64\,$\pm$\,0.03 & 46.70\,$\pm$\,0.09 & 45.64\,$\pm$\,1.12 & 25.94\,$\pm$\,0.47 & 48.66 & 49.06 \\
 & LearnThresh & 644 & 13.52\,$\pm$\,0.01 & 57.43\,$\pm$\,0.34 & 50.40\,$\pm$\,0.34 & 65.90\,$\pm$\,0.07 & 46.67\,$\pm$\,0.23 & \textbf{46.10\,$\pm$\,1.18} & \textbf{26.99\,$\pm$\,1.12} & \textbf{48.91} & \textbf{49.36} \\
\midrule
\multirow{3}{*}{7B (2s)}
 & QLoRA       & 19712 & 10.39\,$\pm$\,0.08 & 70.47\,$\pm$\,0.23 & 61.48\,$\pm$\,0.64 & 79.53\,$\pm$\,0.09 & 59.70\,$\pm$\,0.14 & 67.70\,$\pm$\,0.76 & 43.34\,$\pm$\,1.02 & 63.70 & 64.50 \\
 & NormScale   & \textbf{1303} & 9.87\,$\pm$\,0.21 & \textbf{72.40\,$\pm$\,0.04} & 64.08\,$\pm$\,0.25 & 79.52\,$\pm$\,0.03 & \textbf{60.77\,$\pm$\,0.39} & 70.44\,$\pm$\,1.22 & 51.48\,$\pm$\,0.28 & 66.45 & 67.58 \\
 & LearnThresh & 1344 & \textbf{9.86\,$\pm$\,0.15} & 72.34\,$\pm$\,0.08 & \textbf{64.29\,$\pm$\,0.64} & \textbf{79.70\,$\pm$\,0.06} & 60.28\,$\pm$\,0.06 & \textbf{71.00\,$\pm$\,0.95} & \textbf{52.68\,$\pm$\,0.55} & \textbf{66.72} & \textbf{68.00} \\
\midrule
\multirow{3}{*}{72B (3s)}
 & QLoRA       & 128000 & 5.54\,$\pm$\,0.02 & 83.65\,$\pm$\,0.00 & 71.02\,$\pm$\,0.68 & 84.76\,$\pm$\,0.17 & 62.08\,$\pm$\,0.23 & 85.14\,$\pm$\,0.40 & 52.99\,$\pm$\,1.23 & 73.27 & 75.51 \\
 & NormScale   & \textbf{8205}  & 5.48\,$\pm$\,0.01 & 83.73\,$\pm$\,0.07 & 72.98\,$\pm$\,0.27 & \textbf{86.67\,$\pm$\,0.11} & 64.19\,$\pm$\,1.65 & 88.76\,$\pm$\,0.83 & \textbf{61.43\,$\pm$\,1.68} & 76.29 & \textbf{78.71} \\
 & LearnThresh & 8323  & \textbf{5.45\,$\pm$\,0.01} & \textbf{83.74\,$\pm$\,0.14} & \textbf{73.21\,$\pm$\,0.09} & 86.59\,$\pm$\,0.10 & \textbf{64.53\,$\pm$\,0.50} & \textbf{88.93\,$\pm$\,0.62} & 60.88\,$\pm$\,1.05 & \textbf{76.31} & 78.67 \\
\bottomrule
\end{tabular}}
\end{table}

Table~\ref{tab:main-qwen} shows BQ-LoRA wins or matches QLoRA across all three Qwen2.5
scales.
On 7B (2 seeds), NormScale and LearnThresh lift Avg from $63.70$ to $66.45$/$66.72$
($\sim 3$\,pts) and Avg$^*$ from $64.50$ to $67.58$/$68.00$ ($\sim 3.5$\,pts) at
$15\times$ smaller storage; PPL is preserved within $0.06$ of the untuned base while
QLoRA worsens it by $0.59$.
The single largest contributor is IFEval, where QLoRA loses $11$\,pts vs.\ the untuned
Instruct base ($54.34 \to 43.34$) while LearnThresh loses only $1.7$\,pts ($52.68$);
i.e.\ BQ adapts to Alpaca without forgetting the Instruct model's instruction-following
ability.
On 1.5B (3 seeds), the same pattern holds with smaller margins: BQ wins both averages
($+0.8$--$1.0$\,Avg, $+1.3$--$1.6$\,Avg$^*$) at $13$--$14\times$ smaller storage. The smaller margin
reflects less instruction-following headroom in the smaller base
($27.17$ IFEval vs.\ $54.34$ for 7B).
On 72B (3 seeds), the pattern persists at scale: BQ leads on five of six benchmarks and
ties on MMLU ($\Delta \leq 0.09$), with the largest margins on GSM8K ($+3.6$--$3.8$\,pts) and
IFEval, which again shows the same forgetting asymmetry ($+7.9$--$8.4$\,pts vs.\ QLoRA).
Every other gap exceeds three times the pooled seed standard deviation, except NormScale's
TQA, where one seed is an outlier ($65.14 \pm 0.05$ without it).
WikiText-2 PPL also improves modestly ($-0.06$ for NormScale, $-0.09$ for LearnThresh).

\subsection{SOTA comparison: LLaMA-2}

\begin{table}[t]
\centering
\caption{LLaMA-2 SOTA. 7B: 3 seeds, mean$\pm$std. 70B: 1 seed.
Avg = mean of all six listed benchmarks; Avg$^*$ excludes TQA-MC2.}
\label{tab:sota-llama}
\small
\adjustbox{max width=\textwidth}{%
\begin{tabular}{llcccccccc}
\toprule
Model & Method & MMLU & ARC-C & Hella & TQA-MC2 & GSM8K & IFEval & Avg & Avg$^*$ \\
\midrule
\multirow{4}{*}{7B}
 & QLoRA       & 41.77\,$\pm$\,0.58 & 54.58\,$\pm$\,0.28 & 78.76\,$\pm$\,0.34 & \textbf{52.31\,$\pm$\,1.20} & 10.61\,$\pm$\,0.71 & 17.50\,$\pm$\,0.83 & 42.59 & 40.64 \\
 & LoftQ       & 43.63\,$\pm$\,0.36 & \textbf{55.40\,$\pm$\,0.31} & \textbf{79.20\,$\pm$\,0.16} & 50.99\,$\pm$\,0.25 & 11.85\,$\pm$\,0.14 & 17.81\,$\pm$\,0.63 & \textbf{43.15} & \textbf{41.58} \\
 & NormScale   & 44.11\,$\pm$\,0.11 & 52.90\,$\pm$\,0.07 & 78.01\,$\pm$\,0.05 & 43.69\,$\pm$\,0.33 & 13.24\,$\pm$\,0.31 & \textbf{18.91\,$\pm$\,0.53} & 41.81 & 41.43 \\
 & LearnThresh & \textbf{44.19\,$\pm$\,0.20} & 52.98\,$\pm$\,0.39 & 77.96\,$\pm$\,0.03 & 43.46\,$\pm$\,0.35 & \textbf{13.34\,$\pm$\,0.06} & 18.24\,$\pm$\,0.53 & 41.70 & 41.34 \\
\midrule
\multirow{4}{*}{70B}
 & QLoRA       & \textbf{68.41} & \textbf{68.26} & \textbf{86.62} & \textbf{60.13} & 49.51 & 22.37 & \textbf{59.22} & \textbf{59.03} \\
 & LoftQ       & 67.61          & 67.92          & 86.55          & 58.78          & 48.07 & \textbf{23.11} & 58.67 & 58.65 \\
 & NormScale   & 67.20          & 67.24          & 86.12          & 52.50          & 49.36 & 20.52 & 57.16 & 58.09 \\
 & LearnThresh & 66.64          & 67.49          & 86.27          & 57.39          & \textbf{50.04} & 20.52 & 58.06 & 58.19 \\
\bottomrule
\end{tabular}}
\end{table}

LLaMA-2 results (Table~\ref{tab:sota-llama}) tell a more nuanced story than Qwen.
On 7B, LoftQ leads on Avg ($43.15$) and Avg$^*$ ($41.58$); BQ variants close most of
the gap on Avg$^*$ ($41.34$--$41.43$, within $0.25$\,pt of LoftQ) but pay a $\sim 9$\,pt
TQA-MC2 penalty that drags Avg down.
On 70B, QLoRA leads narrowly ($\sim 1$\,pt on Avg, $0.84$\,pt on Avg$^*$) and BQ matches
LoftQ within $0.6$\,pts. The differentiator at this scale is storage: BQ matches at
$16\times$ smaller adapter footprint, the operating point that matters for many-adapter
serving (\S\ref{sec:sgmv}).
Appendix~\ref{app:geometry} suggests why: BQ updates are near-isotropic on both families,
but LLaMA-2 drives $1.65\times$ larger gates, consistent with its less extensively
pretrained base needing more new behavior, which concentrated FP16 directions supply more
directly.

\subsection{BQ-DyLoRA}
\label{sec:bqdylora}

\begin{table}[t]
\centering
\caption{BQ-DyLoRA vs.\ QDyLoRA on Qwen2.5-7B-Instruct, 3 seeds (42/123/7), mean$\pm$std.
rank$^\ast$ is selected per seed from $\{1,2,4,8,16\}$ by WikiText-2 validation PPL, and KB
is at the selected rank. PPL (WikiText-2 test) is seed 42 only.
Avg = mean of 5 listed task metrics; Avg$^*$ excludes TQA.}
\label{tab:bqdylora}
\small
\adjustbox{max width=\textwidth}{%
\begin{tabular}{lrrcccccccc}
\toprule
Method & KB & rank$^\ast$ & PPL$\downarrow$ & MMLU & ARC-C & Hella & TQA & GSM8K & Avg & Avg$^*$ \\
\midrule
QDyLoRA   & 4928                & 4/4/4   & 9.79          & \textbf{73.05\,$\pm$\,0.19} & 62.20\,$\pm$\,0.88 & \textbf{80.42\,$\pm$\,0.37} & \textbf{62.13\,$\pm$\,0.20} & 70.38\,$\pm$\,4.16 & 69.64 & 71.51 \\
BQ-DyLoRA & \textbf{379--1303}  & 4/16/16 & \textbf{9.58} & 72.59\,$\pm$\,0.32 & \textbf{63.48\,$\pm$\,1.62} & 80.15\,$\pm$\,0.30 & 60.74\,$\pm$\,0.99 & \textbf{73.24\,$\pm$\,0.72} & \textbf{70.04} & \textbf{72.36} \\
\bottomrule
\end{tabular}}
\end{table}

Applying NormScale to QDyLoRA~\citep{rajabzadeh2024qdylora} yields BQ-DyLoRA. Across three
seeds, BQ-DyLoRA leads on ARC-C ($+1.3$) and GSM8K ($+2.9$) and trails by at most $1.4$
points elsewhere (MMLU $-0.5$, HellaSwag $-0.3$, TQA $-1.4$), raising Avg from $69.64$ to
$70.04$ and Avg$^*$ from $71.51$ to $72.36$; the GSM8K gain lies within QDyLoRA's seed
spread ($\pm 4.2$), so we read it only directionally. On seed 42, PPL drops from $9.79$ to
$9.58$, below the $9.80$ untuned base.
The selected ranks differ: QDyLoRA selects $r^\ast{=}4$ on all three seeds, while
BQ-DyLoRA keeps the full $r^\ast{=}16$ on two of three, so the stored adapter is
$3.8$--$13\times$ smaller ($379$--$1{,}303$\,KB vs.\ $4{,}928$\,KB) depending on the seed.

\section{The Binary Constraint as Implicit Regularizer}
\label{sec:reg}

The theory in \S\ref{sec:theory} predicts that BQ-LoRA should under-fit training relative
to QLoRA but generalize at least as well, because the binary parameterization controls
norm and rank \emph{architecturally}, not via weight decay. We test this prediction in
two complementary ways: a dataset ablation isolating the regularization-vs-heterogeneity
relationship, and a direct comparison of training-loss curves to test perplexity.

\subsection{Dataset ablation: gap widens with task heterogeneity}

\begin{table}[t]
\centering
\caption{Dataset ablation on Qwen2.5-7B-Instruct (3000 steps). OpenHermes task metrics are
mean$\pm$std over 3 seeds (PPL: seed 42); Mixed is single-seed.
Avg = mean of 5 listed task metrics; Avg$^*$ excludes TQA.}
\label{tab:dataset}
\small
\adjustbox{max width=\textwidth}{%
\begin{tabular}{llrccccccccc}
\toprule
Dataset & Method & KB & PPL$\downarrow$ & MMLU & ARC-C & Hella & TQA & GSM8K & Avg & Avg$^*$ \\
\midrule
\multirow{3}{*}{OpenHermes}
 & QLoRA       & 19712 & 10.16 & 71.87\,$\pm$\,0.30 & 61.32\,$\pm$\,0.57 & \textbf{79.70\,$\pm$\,0.15} & 56.58\,$\pm$\,0.35 & \textbf{70.71\,$\pm$\,2.42} & 68.04 & 70.90 \\
 & NormScale   & 1303  & \textbf{10.10} & 72.28\,$\pm$\,0.05 & \textbf{62.68\,$\pm$\,0.43} & 79.18\,$\pm$\,0.05 & 58.50\,$\pm$\,1.09 & 70.26\,$\pm$\,0.19 & 68.58 & \textbf{71.10} \\
 & LearnThresh & 1344  & \textbf{10.10} & \textbf{72.38\,$\pm$\,0.10} & 62.49\,$\pm$\,0.82 & 79.24\,$\pm$\,0.08 & \textbf{59.44\,$\pm$\,0.60} & 69.78\,$\pm$\,0.46 & \textbf{68.67} & 70.97 \\
\midrule
\multirow{3}{*}{Mixed}
 & QLoRA       & 19712 & 10.26 & 71.37 & 61.69 & \textbf{79.88} & 58.09 & 60.20 & 66.25 & 68.28 \\
 & NormScale   & 1303  & \textbf{10.00} & 72.48 & 63.99 & 79.29 & 58.42 & \textbf{74.53} & \textbf{69.74} & \textbf{72.57} \\
 & LearnThresh & 1344  & 10.05 & \textbf{72.62} & \textbf{64.51} & 79.38 & \textbf{58.83} & 73.09 & 69.69 & 72.40 \\
\bottomrule
\end{tabular}}
\end{table}

OpenHermes is curated and stylistically consistent; Mixed combines 10k Alpaca and 10k
OpenHermes samples, two instruction sources with different styles. The regularization story predicts the BQ--QLoRA gap should be
\emph{larger} on Mixed because more heterogeneous data offers more opportunities to
overfit individual styles. Table~\ref{tab:dataset} confirms this: on OpenHermes (3 seeds)
the Avg gap is $\sim 0.6$\,pt; on Mixed it is $\sim 3.5$\,pts (NormScale $69.74$ vs.\
QLoRA $66.25$), driven primarily by GSM8K ($+14$\,pts absolute).

\subsection{Loss projection: BQ under-fits training, matches or beats on test}

\begin{figure}[t]
\centering
\includegraphics[width=0.85\textwidth]{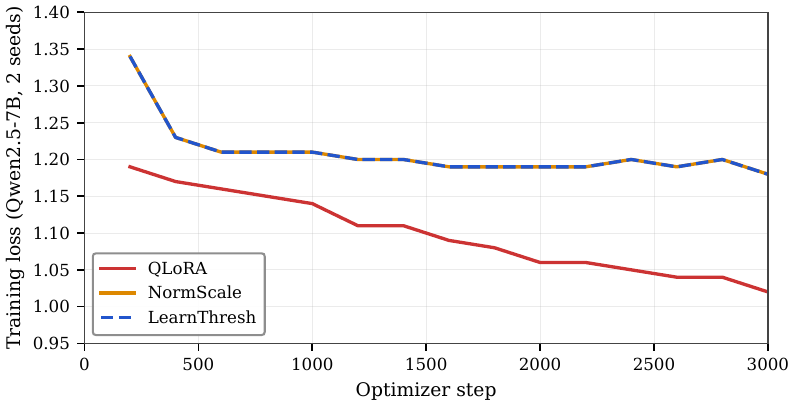}
\caption{Training loss on Qwen2.5-7B (Alpaca, 3{,}000 steps, mean of 2 seeds).
QLoRA descends to $\approx 1.02$; BQ variants plateau at $\approx 1.18$, a $0.16$-nat
gap. Test perplexity, however, is \emph{better} for the BQ variants ($9.86$ vs.\
$10.39$, Table~\ref{tab:main-qwen}), and benchmark Avg is $\sim 3$\,pts higher.
NormScale and LearnThresh trajectories are point-wise identical, consistent with their
shared hypothesis class up to a per-layer scalar threshold.}
\label{fig:trainloss}
\end{figure}

Figure~\ref{fig:trainloss} plots training-loss trajectories on Qwen2.5-7B.
QLoRA drives train loss to $\approx 1.02$; BQ variants plateau at $\approx 1.18$, a
$0.16$-nat gap. On test perplexity the order \emph{reverses}---BQ wins by $0.53$\,PPL.
This is the canonical regularization signature: the constrained model fits training data
less aggressively but generalizes better, exactly as the hypothesis-class size argument
of \S\ref{sec:theory} predicts.
We replicate this pattern at three additional scales (Qwen2.5-1.5B, Qwen2.5-72B,
LLaMA-2-7B) in Appendix~\ref{app:loss-curves}.

\section{BQ-SGMV: Many-Adapter Serving}
\label{sec:sgmv}

We extend Punica's SGMV~\citep{chen2024punica} with a binary inner-product primitive,
prove a cache phase-transition result, and validate on A100-80GB with LLaMA-2-7B.
\textbf{Scope.} All serving numbers here come from our research prototype under a synthetic
Zipfian workload. What transfers to any serving stack is the cache-capacity arithmetic
($\approx 15\times$ more adapters per GB); throughput and tail latency are claimed only for
this prototype, and vLLM/TGI integration is future work.

\subsection{Existing SGMV and BQ-SGMV}
SGMV (Segmented Gather Matrix-Vector) batches adapter forward passes for many
concurrent requests by gathering $A, B$ slices per request group and performing two
FP16 GEMVs. We replace the FP16 inner products with a packed-binary inner product
(popcount-based, following XNOR-Net~\citep{rastegari2016xnor}) and a single scale
multiply per output tile. The change is local to the kernel.

\begin{figure}[h]
\centering
\begin{minipage}[t]{0.48\textwidth}
\begin{algorithm}[H]
\caption{SGMV (Punica)}\label{alg:sgmv}
\begin{algorithmic}[1]
\State \textbf{Input:} $X \in \mathbb{R}^{B \times d_{\text{in}}}$, $\mathrm{adapter\_id}$
\State \textbf{Output:} $Y \in \mathbb{R}^{B \times d_{\text{out}}}$
\For{group $g$ in batch}
  \State $A \gets A_{\text{pool}}[\mathrm{adapter\_id}_g]$
  \Statex \quad\textit{// FP16, $r \times d_{\text{in}}$}
  \State $H \gets X_g\, A^\top$ \Comment{FP16 GEMV}
  \State $B \gets B_{\text{pool}}[\mathrm{adapter\_id}_g]$
  \Statex \quad\textit{// FP16, $d_{\text{out}} \times r$}
  \State $Y_g \mathrel{+}= H\, B^\top$ \Comment{FP16 GEMV}
\EndFor
\end{algorithmic}
\end{algorithm}
\end{minipage}\hfill
\begin{minipage}[t]{0.48\textwidth}
\begin{algorithm}[H]
\caption{BQ-SGMV (ours)}\label{alg:bqsgmv}
\begin{algorithmic}[1]
\State \textbf{Input:} $X$, $\mathrm{adapter\_id}$
\State \textbf{Output:} $Y$
\For{group $g$ in batch}
  \State $\tilde{A} \gets A_{\text{bitpool}}[\mathrm{adapter\_id}_g]$
  \Statex \quad\textit{// 1-bit packed}
  \State $H \gets \texttt{binary\_gemv}(X_g,\, \tilde{A})$
  \State $\tilde{B} \gets B_{\text{bitpool}}[\mathrm{adapter\_id}_g]$
  \State $s \gets s_{\text{pool}}[\mathrm{adapter\_id}_g]$
  \State $Y_g \mathrel{+}= s \cdot \texttt{binary\_gemv}(H,\, \tilde{B})$
\EndFor
\end{algorithmic}
\end{algorithm}
\end{minipage}
\end{figure}

The change to memory layout is dominant: $A_{\text{pool}}$ shrinks $16\times$, allowing
$16\times$ more adapters in HBM at fixed budget.

\subsection{Cache phase transition}

\begin{proposition}[Cache phase transition]
\label{prop:cache}
For an LRU adapter cache of capacity $C$ adapters under independent Zipfian requests with
exponent $\theta > 1$ over a working set of $N$ adapters, the steady-state miss rate
(excluding compulsory misses) satisfies
\begin{equation}
  \mu(C, N, \theta) \;=\;
  \begin{cases}
    0, & C \geq N, \\
    \Theta\!\left(\dfrac{C^{1-\theta} - N^{1-\theta}}{1 - N^{1-\theta}}\right), & C < N.
  \end{cases}
\end{equation}
Increasing the cache from $C_{\text{QL}}$ to $C_{\text{BQ}} = \lambda C_{\text{QL}}$
($\lambda \approx 16$) gives:
(i) for $N \gg C_{\text{BQ}}$, a miss-rate reduction by a factor $\Theta(\lambda^{\theta-1})$,
and a larger one as $C_{\text{BQ}}$ approaches $N$;
(ii) a phase transition to zero misses when $C_{\text{QL}} < N \leq C_{\text{BQ}}$.
\end{proposition}
\begin{proof}[Sketch]
Under the Che approximation~\citep{fricker2012versatile}, an LRU cache of capacity $C$
behaves, up to constants depending only on $\theta$, like a cache holding the $\Theta(C)$
most popular items, so $\mu$ is of the order of the Zipf tail mass beyond rank $C$:
$\sum_{k=C+1}^{N} k^{-\theta} \big/ \sum_{k=1}^{N} k^{-\theta} \asymp
(C^{1-\theta} - N^{1-\theta})/(1 - N^{1-\theta})$. As $N \to \infty$ this is
$\Theta(C^{1-\theta})$, which gives (i). (ii) is immediate: $C \geq N$ implies every
adapter fits.
\end{proof}

\begin{remark}[Throughput consequence]
Per-request time $L = L_{\text{base}} + p L_{\text{transfer}} + L_{\text{compute}}$
where $p$ is miss probability. Replacing $L_{\text{transfer}}^{\text{QL}} = 2.7$\,ms
with $L_{\text{transfer}}^{\text{BQ}} = 0.19$\,ms and reducing $p$ via
Proposition~\ref{prop:cache} multiplies these effects.
\end{remark}

\subsection{Empirical validation}

\begin{table}[t]
\centering
\caption{End-to-end serving on LLaMA-2-7B (A100-80GB, Zipfian $\theta = 1.2$).}
\label{tab:sgmv}
\small
\adjustbox{max width=\textwidth}{%
\begin{tabular}{lllrrrrr}
\toprule
Scenario & Method & Cache (\#adapt.) & rps & miss\% & swap (ms) & p99 (ms) \\
\midrule
small    & QLoRA   & 1{,}250    & 570.0 & 0.0\%  & 0.00 & 2.1 \\
small    & BQ-SGMV & 19{,}500   & \textbf{742.4} & 0.0\%  & 0.00 & \textbf{1.4} \\
medium   & QLoRA   & 1{,}250    & 741.2 & 0.0\%  & 0.00 & 1.5 \\
medium   & BQ-SGMV & 19{,}500   & 705.3 & 0.0\%  & 0.00 & 2.0 \\
pressure & QLoRA   & 250        & 518.4 & 13.0\% & 4.23 & 5.7 \\
pressure & BQ-SGMV & 3{,}900    & \textbf{717.6} & \textbf{0.0\%}  & \textbf{0.00} & \textbf{1.6} \\
\bottomrule
\end{tabular}}
\end{table}

\textbf{(A) PCIe swap.} QLoRA: $2.745 \pm 0.031$\,ms ($12.2$\,GB/s); BQ-LoRA:
$0.189 \pm 0.002$\,ms ($16\times$ smaller).
\textbf{(B) HBM capacity.} On a single A100-80GB with LLaMA-2-7B base, QLoRA admits
$2{,}500$ co-loaded adapters; BQ-SGMV admits $39{,}000$ ($15.6\times$).
\textbf{(C) End-to-end} (Table~\ref{tab:sgmv}). Under cache pressure
($N \approx 1{,}000 > C_{\text{QL}} = 250$), QLoRA misses $13\%$ and rps drops to $518$.
BQ-SGMV ($C_{\text{BQ}} = 3{,}900 \geq N$) is in the phase-transition regime of
Proposition~\ref{prop:cache}, sustains $0\%$ misses, and improves throughput
$1.4\times$ and tail latency $3.6\times$.

\section{Discussion}
\label{sec:discussion}
\paragraph{Why we report Avg and Avg$^*$.}
On Qwen2.5, BQ-LoRA wins or ties on PPL, MMLU, ARC-C, GSM8K, and IFEval at every scale;
on LLaMA-2 the picture is mixed (Table~\ref{tab:sota-llama}).
Its behavior on TQA-MC2 is non-monotonic with the base model:
BQ wins on Qwen2.5-7B and Qwen2.5-72B ($+0.6$--$1.1$ and $+2.1$--$2.5$\,pts vs.\ QLoRA),
is slightly behind on Qwen2.5-1.5B ($-1.6$\,pts), and loses on LLaMA-2
($\approx -9$\,pts at 7B; $-3$ to $-8$\,pts at 70B).
HellaSwag follows a similar pattern.
We report two averages to make this transparent. Avg includes TQA-MC2; Avg$^*$ excludes
it. Both are reported in every main table; we do not claim the choice of average should
drive the reading.

\paragraph{Where this leaves the PEFT design space.}
The standard view of LoRA is that the rank-$r$ structure is the inductive bias and FP16
is just a precision choice. Our results suggest the opposite framing is at least as
useful: the bit-width is itself an inductive bias, and lowering it to one bit gives
explicit, architectural norm control that rank alone does not; 2- and 4-bit adapters do
not beat 1 bit (Appendix~\ref{app:bitwidth}). The same principle could
be applied to other PEFT parameterizations (prefix tuning~\citep{li2021prefix},
$\text{IA}^3$~\citep{liu2022ia3}, AdaLoRA~\citep{zhang2023adalora}); we view BQ-LoRA as
the simplest demonstration of the idea.

\paragraph{Limitations.}
We use only $\{q,k,v,o\}$ adapter targets at rank 16; whether the regularization gap
persists for adapters on MLP projections or at much higher ranks is untested.
The LLaMA-2-70B results and the Mixed dataset ablation are single-seed; we prioritized
Qwen2.5-72B as the large-scale multi-seed setting.
Theorem~\ref{thm:gen} bounds a single adapted layer acting linearly on its input; its
typical-set form relies on trained signs staying random-like, which we verify
empirically (Appendix~\ref{app:sign-spectra}) rather than prove, and tighter data-dependent bounds in the spirit of \citet{nagarajan2019uniform} or
PAC-Bayes treatments of low-rank fine-tuning remain open.
A version-control lapse during development meant that not all runs used identical library
versions; re-running representative Qwen2.5-7B configurations under pinned versions showed
the effect on our conclusions to be negligible.
The BQ-SGMV kernel (Algorithm~\ref{alg:bqsgmv}) is a research prototype evaluated on a
synthetic workload; integration with production serving systems such as vLLM or TGI is
future work.
Finally, we evaluate only English instruction-tuning data; whether the regularization
gap transfers to multilingual or code-heavy fine-tuning is open.

\section{Conclusion}
We showed that the FP16 adapter precision used in QLoRA-style fine-tuning of quantized
language models is largely unnecessary. A single bit per adapter parameter, gated by a
learned per-layer scale and trained with a Scaled Straight-Through Estimator, matches or
exceeds QLoRA on benchmark averages at every Qwen2.5 scale while shrinking inference-time
adapter storage by $13$--$16\times$ at essentially unchanged training cost; on LLaMA-2 it
remains competitive but does not uniformly lead the strongest FP16-adapter baselines.
The binary constraint acts as an architectural prior: the hypothesis class has bounded
norm, full rank and near-isotropy, and an explicit Rademacher bound absent from the FP16
case, and the predicted signature (under-fitting on training, matching or beating on
test, a heterogeneity-dependent gap) appears in our ablations and loss curves. One
A100-80GB holds $15.6\times$ more adapters, which provably removes cache misses whenever
the working set fits. Open directions include adapters beyond $\{q,k,v,o\}$, the TQA-MC2
gap on LLaMA-2, and natively low-bit bases such as BitNet b1.58~\citep{ma2024bitnet158}.

\bibliographystyle{plainnat}

\appendix

\section{Related Work}
\label{app:related}

\paragraph{Parameter-efficient fine-tuning (PEFT).}
The original adapter formulation~\citep{houlsby2019adapter} inserts small bottleneck
layers in transformer blocks; subsequent work explored alternative parameterizations
including prefix tuning~\citep{li2021prefix}, prompt tuning~\citep{lester2021prompt},
$\text{IA}^3$~\citep{liu2022ia3}, LoRA~\citep{hu2021lora}, and rank-adaptive
schemes such as AdaLoRA~\citep{zhang2023adalora} and DyLoRA~\citep{valipour2022dylora}.
LoRA emerged as the dominant variant on cost grounds: it adds zero inference latency
when merged and trains a small fraction of parameters. BQ-LoRA inherits LoRA's structure
but replaces FP16 with one bit, which we view as orthogonal: the same bit-width
reduction could in principle apply to prefix or $\text{IA}^3$ parameters.

\paragraph{Post-training and quantization-aware fine-tuning.}
QLoRA~\citep{dettmers2023qlora} pairs 4-bit NF4 base with FP16 LoRA adapters.
LoftQ~\citep{li2023loftq} initializes the adapter to compensate for base-weight
quantization error via alternating SVD. QDyLoRA~\citep{rajabzadeh2024qdylora}
extends QLoRA to nested ranks via rank dropout, building on
DyLoRA~\citep{valipour2022dylora}. All three keep the adapter in FP16/BF16.
Pure post-training base quantization---without adapters---is studied in
GPTQ~\citep{frantar2023gptq} and AWQ~\citep{lin2024awq}, both of which complement BQ-LoRA
on the base side: any of NF4, GPTQ, or AWQ can be used as the frozen base under our
binary adapter.
Recent work pushes adapter precision further: LowRA~\citep{zhou2025lowra} achieves
LoRA fine-tuning below 2 bits per parameter via fine-grained quantization mapping;
Q-BLoRA~\citep{shen2025qblora} re-balances adapter rank and input/output complexity to
mitigate underfitting; CoA-LoRA~\citep{ye2025coalora} adapts a single LoRA across
heterogeneous quantization configurations.
BitDelta~\citep{liu2024bitdelta} instead binarizes the full-rank weight delta of a fully
fine-tuned model; its storage therefore scales with model size ($\approx 0.8$\,GB for a 7B
model, versus $1.3$\,MB for a rank-16 BQ-LoRA adapter on Qwen2.5-7B), so the two operate
at storage budgets nearly three orders of magnitude apart.
BQ-LoRA reduces the adapter to one bit per parameter plus a single scale, with
explicit theoretical norm control absent in the prior work.

\paragraph{Binary, ternary, and quantized networks.}
Binary~\citep{courbariaux2016binarynet,rastegari2016xnor} and
ternary~\citep{li2016ternary} networks have a long history; the
quantized-neural-network framework of \citet{hubara2017qnn} unified low-precision
weights and activations under STE-based training. Recent BitNet~\citep{wang2023bitnet}
and BitNet b1.58~\citep{ma2024bitnet158} show extreme weight quantization can scale to
LLMs when models are trained from scratch in low precision. BQ-LoRA is complementary:
it leaves the pretrained 4-bit base intact and binarizes only the small adapter,
enabling post-hoc fine-tuning of any existing checkpoint without from-scratch training.

\paragraph{Multi-adapter serving.}
S-LoRA~\citep{sheng2023slora}, Punica~\citep{chen2024punica}, and
dLoRA~\citep{wu2024dlora} optimize many-adapter inference via custom SGMV kernels and
unified memory paging. The fundamental scarce resource is HBM: cache hit rate is
bounded by the number of adapters that fit. By shrinking the per-adapter footprint,
BQ-LoRA directly relieves this bound.

\paragraph{Implicit regularization and generalization theory.}
\citet{biderman2024lora} and \citet{zhu2024asymmetry} attribute LoRA's generalization
properties to its inductive bias. Classical norm-based generalization results we
build on include the spectral margin bound of \citet{bartlett2017spectrally}, the
vector-contraction Rademacher bound of \citet{maurer2016vector}, and standard text
treatments~\citep{mohri2018foundations}. \citet{neyshabur2017exploring} survey
norm-based generalization measures in deep networks; \citet{nagarajan2019uniform}
caution that uniform-convergence bounds may be vacuous in over-parameterized regimes,
which motivates our choice of an architectural (parameterization-level) constraint
rather than a purely norm-based one.

\paragraph{Base models and instruction data.}
Our evaluations use Qwen2.5-Instruct~\citep{qwen2024qwen25} and
LLaMA-2~\citep{touvron2023llama2} as 4-bit-quantized bases, fine-tuned on
Alpaca~\citep{taori2023alpaca} (the standard PEFT comparison dataset); the dataset
ablation (Table~\ref{tab:dataset}) additionally uses OpenHermes alone and a
heterogeneity-stress mixture (10k Alpaca + 10k OpenHermes-2.5).

\section{Worst-case Frobenius bound}
\label{app:wc-frob}

The worst-case Frobenius bound complementing Proposition~\ref{prop:hp-norm} can be
derived without high-probability tools. For any
$\tilde{A} \in \{\pm 1\}^{r \times d_{\text{in}}}, \tilde{B} \in \{\pm 1\}^{d_{\text{out}} \times r}$:
\begin{align*}
\|\tilde{B}\tilde{A}\|_F
\;\leq\; \|\tilde{B}\|_F\, \|\tilde{A}\|_2
\;\leq\; \|\tilde{B}\|_F\, \|\tilde{A}\|_F
\;=\; \sqrt{r d_{\text{out}}} \cdot \sqrt{r d_{\text{in}}}
\;=\; r\, \sqrt{d_{\text{in}} d_{\text{out}}}.
\end{align*}
The bound is attained by constant sign matrices
($\tilde{B}\tilde{A} = r\,\mathbf{1}\mathbf{1}^\top$), whose spectral norm is also
$r\sqrt{d_{\text{in}} d_{\text{out}}}$; multiplied by $s_{\max}\gamma$ it gives the worst case
of Proposition~\ref{prop:hp-norm}(a). It exceeds the typical-set bound of
Proposition~\ref{prop:hp-norm}(b) by a factor $\Theta(r)$: for random signs both
$\|\tilde{A}\|_F = \sqrt{r d_{\text{in}}}$ and $\|\tilde{B}\|_F = \sqrt{r d_{\text{out}}}$
exceed the corresponding spectral norms, $\Theta(\sqrt{r} + \sqrt{d_{\text{in}}})$ and
$\Theta(\sqrt{r} + \sqrt{d_{\text{out}}})$, by a factor of about $\sqrt{r}$ each
(Lemma~\ref{lem:opnorm}).

\paragraph{Uniform finite-class bound.}
\label{app:massart}
The sign part of $\mathcal{H}_{\text{BQ}}$ is a finite set
$\mathcal{S} = \{\gamma\tilde{B}\tilde{A}\}$ with
$|\mathcal{S}| \leq 2^{r(d_{\text{in}} + d_{\text{out}})}$, closed under negation. Since
$\sup_{|s| \leq s_{\max}} s Z = s_{\max}|Z|$, the empirical Rademacher complexity of
$\mathcal{H}_{\text{BQ}}(s_{\max})$ in the setting of Theorem~\ref{thm:gen} is $s_{\max}$
times that of $\mathcal{S}$. Every member satisfies $|u^\top W x| \leq \alpha\sqrt{r\,
d_{\text{out}}}\,R$ (Proposition~\ref{prop:hp-norm}(a) with $s = 1$), so Massart's
finite-class lemma~\citep{mohri2018foundations} gives
\begin{equation*}
\hat{\mathfrak{R}}_n \;\leq\; s_{\max}\,\alpha\sqrt{r\, d_{\text{out}}}\; R\,
\sqrt{\frac{2\, r(d_{\text{in}} + d_{\text{out}}) \log 2}{n}}
\end{equation*}
uniformly over the class, trained members included, with no assumption on how the signs
arise; $\log|\mathcal{S}|$ recovers the $r(d_{\text{in}} + d_{\text{out}})$ capacity of
the binary parameterization. This bound is looser in the dimensions than
Theorem~\ref{thm:gen}; its role is to show that the class is small without reference to
any distribution over signs.

\section{Spectra of trained sign matrices}
\label{app:sign-spectra}

Proposition~\ref{prop:hp-norm}(c) concerns the signs at initialization. To check whether
training moves them out of the typical set $\mathcal{T}_\delta$, we compare the spectral
norm of every trained sign matrix with a 200-sample Monte-Carlo ensemble of random
$\pm 1$ matrices of the same shape, across four 7B checkpoints (960 matrices;
Table~\ref{tab:sign-spectra}).

\begin{table}[H]
\centering
\caption{Trained sign matrices vs.\ random $\pm 1$ matrices of identical shape. Ratio is
$\sigma_{\max}$ of the trained matrix over the ensemble mean of $\sigma_{\max}$.}
\label{tab:sign-spectra}
\small
\begin{tabular}{lcccc}
\toprule
Checkpoint (\# matrices) & mean ratio & worst ratio & within $2\sigma$ of random & frac$(+1)$ \\
\midrule
Qwen2.5-7B, NormScale (224)   & 1.000 & 1.034 & 94\% & 0.500 \\
Qwen2.5-7B, LearnThresh (224) & 1.000 & 1.035 & 95\% & 0.501 \\
LLaMA-2-7B, NormScale (256)   & 1.000 & 1.014 & 95\% & 0.500 \\
LLaMA-2-7B, LearnThresh (256) & 1.001 & 1.048 & 89\% & 0.501 \\
\bottomrule
\end{tabular}
\end{table}

Every trained sign matrix has spectral norm within $5\%$ of the random ensemble, whose
mean tracks the Bai--Yin value $\sqrt{d} + \sqrt{r}$; matrices outside the $2\sigma$ band
deviate by less than $5\%$ in absolute terms because the ensemble is tightly
concentrated. The fraction of $+1$ entries stays at $0.500$--$0.501$, so the learned
thresholds of LearnThresh remain close to zero and the signs stay balanced. Trained signs
therefore lie in $\mathcal{T}_\delta$ up to a factor $1.05$ in each sign factor, and the
typical-set bounds of Proposition~\ref{prop:hp-norm}(b) and Theorem~\ref{thm:gen} apply
to the trained adapters up to a factor $1.05^2 \approx 1.10$.

\section{Training-Loss Curves at Multiple Scales}
\label{app:loss-curves}

We report training-loss trajectories at four model scales to confirm the regularization
signature of \S\ref{sec:reg} is not a single-scale artifact. The pattern is consistent:
QLoRA descends faster and lower; BQ variants plateau higher; the gap widens with model
size. NormScale and LearnThresh have indistinguishable trajectories at every scale we
tested (the two curves overlap exactly), so we plot a single ``BQ'' curve in figures
and report both rows in tables.

\begin{figure}[H]
\centering
\includegraphics[width=0.95\textwidth]{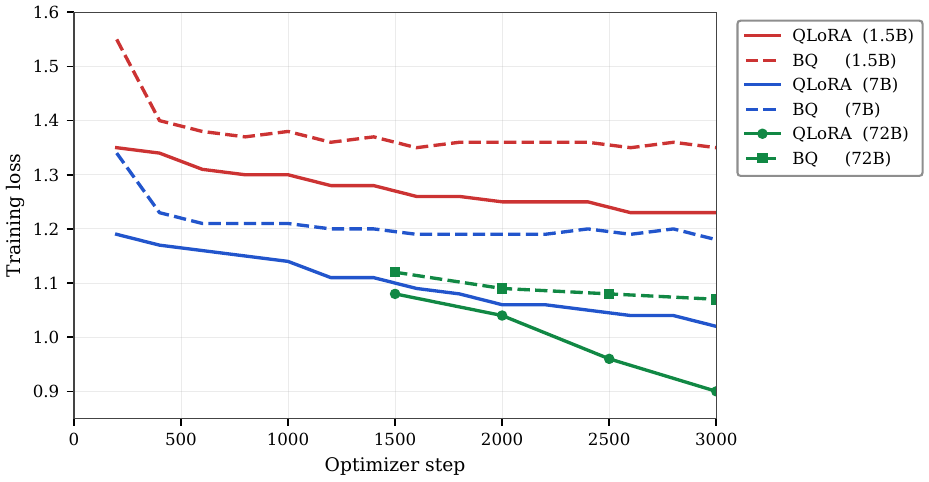}
\caption{Training-loss trajectories on Qwen2.5-Instruct (1.5B / 7B / 72B). BQ plateaus
above QLoRA at every scale; the gap is $\sim 0.12$ nats on 1.5B,
$\sim 0.16$ nats on 7B, and $\sim 0.17$ nats on 72B (final).
NormScale and LearnThresh trajectories are indistinguishable and are plotted as a
single ``BQ'' curve. 72B uses only 4 measured checkpoints (steps 1500/2000/2500/3000).}
\label{fig:qwen-loss}
\end{figure}

\begin{figure}[H]
\centering
\includegraphics[width=0.85\textwidth]{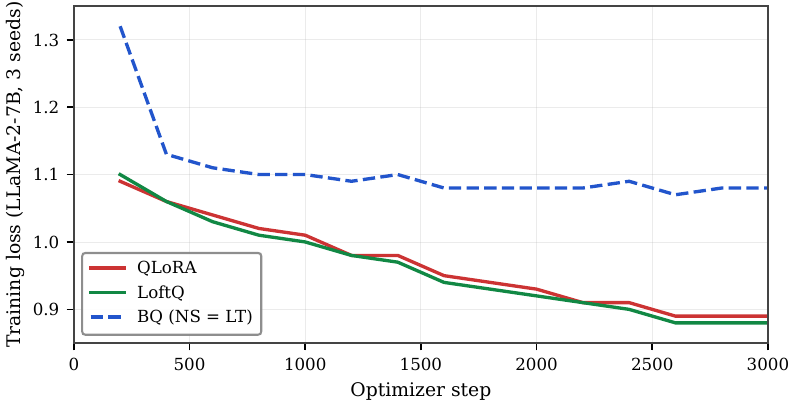}
\caption{LLaMA-2-7B training-loss trajectories (3 seeds). Same pattern as Qwen.}
\label{fig:llama-loss}
\end{figure}

\begin{table}[H]
\centering
\caption{Step-wise training loss on Qwen2.5-1.5B-Instruct (3 seeds, mean$\pm$std).
NormScale and LearnThresh are point-wise identical; we plot one ``BQ'' curve in
Figure~\ref{fig:qwen-loss}.}
\label{tab:qwen15-loss}
\small
\adjustbox{max width=\textwidth}{%
\begin{tabular}{lccccccccc}
\toprule
Method & 200 & 400 & 600 & 800 & 1000 & 1200 & 1400 & 1600 \\
\midrule
QLoRA       & 1.35\,$\pm$\,0.01 & 1.34\,$\pm$\,0.01 & 1.31\,$\pm$\,0.01 & 1.30\,$\pm$\,0.01 & 1.30\,$\pm$\,0.01 & 1.28\,$\pm$\,0.01 & 1.28\,$\pm$\,0.01 & 1.26\,$\pm$\,0.01 \\
NormScale   & 1.55\,$\pm$\,0.01 & 1.40\,$\pm$\,0.01 & 1.38\,$\pm$\,0.01 & 1.37\,$\pm$\,0.01 & 1.38\,$\pm$\,0.01 & 1.36\,$\pm$\,0.01 & 1.37\,$\pm$\,0.01 & 1.35\,$\pm$\,0.01 \\
LearnThresh & 1.55\,$\pm$\,0.01 & 1.40\,$\pm$\,0.01 & 1.38\,$\pm$\,0.01 & 1.37\,$\pm$\,0.01 & 1.38\,$\pm$\,0.01 & 1.36\,$\pm$\,0.01 & 1.37\,$\pm$\,0.01 & 1.35\,$\pm$\,0.01 \\
\midrule
       & 1800 & 2000 & 2200 & 2400 & 2600 & 2800 & 3000 \\
\midrule
QLoRA       & 1.26\,$\pm$\,0.01 & 1.25\,$\pm$\,0.00 & 1.25\,$\pm$\,0.01 & 1.25\,$\pm$\,0.00 & 1.23\,$\pm$\,0.01 & 1.23\,$\pm$\,0.01 & 1.23\,$\pm$\,0.02 \\
NormScale   & 1.36\,$\pm$\,0.01 & 1.36\,$\pm$\,0.00 & 1.36\,$\pm$\,0.01 & 1.36\,$\pm$\,0.01 & 1.35\,$\pm$\,0.01 & 1.36\,$\pm$\,0.01 & 1.35\,$\pm$\,0.02 \\
LearnThresh & 1.36\,$\pm$\,0.01 & 1.36\,$\pm$\,0.00 & 1.36\,$\pm$\,0.01 & 1.36\,$\pm$\,0.01 & 1.35\,$\pm$\,0.01 & 1.36\,$\pm$\,0.02 & 1.35\,$\pm$\,0.02 \\
\bottomrule
\end{tabular}}
\end{table}

\begin{table}[H]
\centering
\caption{Step-wise training loss on Qwen2.5-7B-Instruct (2 seeds, mean$\pm$std).}
\label{tab:qwen7-loss}
\small
\adjustbox{max width=\textwidth}{%
\begin{tabular}{lccccccccc}
\toprule
Method & 200 & 400 & 600 & 800 & 1000 & 1200 & 1400 & 1600 \\
\midrule
QLoRA       & 1.19\,$\pm$\,0.01 & 1.17\,$\pm$\,0.00 & 1.16\,$\pm$\,0.01 & 1.15\,$\pm$\,0.00 & 1.14\,$\pm$\,0.01 & 1.11\,$\pm$\,0.01 & 1.11\,$\pm$\,0.01 & 1.09\,$\pm$\,0.01 \\
NormScale   & 1.34\,$\pm$\,0.01 & 1.23\,$\pm$\,0.00 & 1.21\,$\pm$\,0.01 & 1.21\,$\pm$\,0.00 & 1.21\,$\pm$\,0.01 & 1.20\,$\pm$\,0.01 & 1.20\,$\pm$\,0.01 & 1.19\,$\pm$\,0.01 \\
LearnThresh & 1.34\,$\pm$\,0.01 & 1.23\,$\pm$\,0.00 & 1.21\,$\pm$\,0.01 & 1.21\,$\pm$\,0.00 & 1.21\,$\pm$\,0.01 & 1.20\,$\pm$\,0.01 & 1.20\,$\pm$\,0.01 & 1.19\,$\pm$\,0.01 \\
\midrule
       & 1800 & 2000 & 2200 & 2400 & 2600 & 2800 & 3000 \\
\midrule
QLoRA       & 1.08\,$\pm$\,0.00 & 1.06\,$\pm$\,0.00 & 1.06\,$\pm$\,0.01 & 1.05\,$\pm$\,0.00 & 1.04\,$\pm$\,0.02 & 1.04\,$\pm$\,0.01 & 1.02\,$\pm$\,0.00 \\
NormScale   & 1.19\,$\pm$\,0.01 & 1.19\,$\pm$\,0.00 & 1.19\,$\pm$\,0.01 & 1.20\,$\pm$\,0.00 & 1.19\,$\pm$\,0.02 & 1.20\,$\pm$\,0.01 & 1.18\,$\pm$\,0.00 \\
LearnThresh & 1.19\,$\pm$\,0.01 & 1.19\,$\pm$\,0.00 & 1.19\,$\pm$\,0.01 & 1.20\,$\pm$\,0.00 & 1.19\,$\pm$\,0.02 & 1.20\,$\pm$\,0.01 & 1.18\,$\pm$\,0.00 \\
\bottomrule
\end{tabular}}
\end{table}

\begin{table}[H]
\centering
\caption{Step-wise training loss on Qwen2.5-72B-Instruct (1 seed, sparse logging at
steps 1500 / 2000 / 2500 / 3000).}
\label{tab:qwen72-loss}
\small
\adjustbox{max width=0.55\textwidth}{%
\begin{tabular}{lcccc}
\toprule
Method      & 1500 & 2000 & 2500 & 3000 \\
\midrule
QLoRA       & 1.08 & 1.04 & 0.96 & 0.90 \\
NormScale   & 1.12 & 1.09 & 1.08 & 1.07 \\
LearnThresh & 1.12 & 1.09 & 1.08 & 1.07 \\
\bottomrule
\end{tabular}}
\end{table}

\paragraph{Reading the trajectories.}
Three observations across scales.
First, the BQ--QLoRA training-loss gap exists at every scale: $0.12$ nats on 1.5B,
$0.16$ nats on 7B, $0.17$ nats on 72B (at step 3000), all in the same direction.
Second, NormScale and LearnThresh are point-wise indistinguishable within seed
variance---neither variant fits training better, which is expected since they share
the same hypothesis class up to a per-layer scalar threshold. Third, at 72B the gap is
\emph{larger in absolute terms} but the test-set Avg favors BQ by $3$\,pts
(Table~\ref{tab:main-qwen}); the wider train/test gap at scale is consistent with the
regularization-vs-capacity argument in Theorem~\ref{thm:gen}, where BQ's bounded
hypothesis class becomes increasingly favorable as the unconstrained class grows.

\section{STE Variant Ablation}
\label{app:ste}

We evaluate STE variants on Qwen2.5-7B-Instruct (4-bit NF4, rank 16, 3{,}000 steps,
$\alpha = 32$, single seed). Each variant differs from NormScale in exactly one of
(i) the forward scaling rule, (ii) the STE backward, (iii) the binarization threshold,
or (iv) the initialization of $s$.

\paragraph{Variant definitions.}
All variants share the binary forward
$\Delta W = s\,c\,\mathrm{sign}(B^* - \tau_B)\,\mathrm{sign}(A^* - \tau_A)$ with
$\mathcal{U}(-0.5, 0.5)$ latent initialization and FP32 latent updates; unless stated
otherwise, $c = \gamma = \alpha/\sqrt{r d_{\text{in}}}$, $\tau_A = \tau_B = 0$, and $s_0 = 0$.

\begin{itemize}[leftmargin=*,topsep=0pt,itemsep=2pt]
\item \textbf{NormScale} (main paper). Norm-preserving scaling $c = \gamma$ with the
clipped STE, which replaces the (zero) sub-gradient of $\mathrm{sign}$ by
$\partial\,\mathrm{sign}(x)/\partial x \approx \mathbf{1}[|x| \leq 1]$.

\item \textbf{StdScale}. As NormScale but with the standard LoRA scaling $c = \alpha/r$
in place of $\gamma$.

\item \textbf{ScaledSTE}. As NormScale but with the triangular STE
$\partial\,\mathrm{sign}(x)/\partial x \approx \max(0, 1 - |x|)$.

\item \textbf{LearnThresh} (main paper). As NormScale but with learnable per-matrix
thresholds $\tau_A, \tau_B$, updated jointly with $A^*, B^*, s$; the STE mask is
centered at the threshold, $\mathbf{1}[|x - \tau| \leq 1]$. Allows asymmetric
binarization ($\tilde{A}_{ij} = +1$ when $A^*_{ij} > \tau_A$).

\item \textbf{ScaleInit}$_{0.1}$. As NormScale but with $s_0 = 0.1$ instead of $0$.
Violates the soft-start invariant of Proposition~\ref{prop:soft-start}.
\end{itemize}

\begin{table}[H]
\centering
\caption{STE variants on Qwen2.5-7B-Instruct (single seed, 3{,}000 steps, $\alpha = 32$).
Avg = mean of \{MMLU, ARC-C, HellaSwag, TQA, GSM8K\}; Avg$^*$ excludes TQA.}
\label{tab:ste-ablation}
\small
\adjustbox{max width=\textwidth}{%
\begin{tabular}{lcccccccc}
\toprule
Method            & WT-PPL$\downarrow$ & MMLU & ARC-C & HellaSwag & TQA & GSM8K & Avg & Avg$^*$ \\
\midrule
QLoRA             & 10.47          & 70.70          & 60.84          & 79.43         & 59.56          & 69.45          & 68.00 & 70.11 \\
NormScale         & 9.65           & 72.43          & 63.82          & 79.54         & 61.16          & 72.25          & 69.84 & 72.01 \\
StdScale          & \textbf{9.64}  & 72.31          & 63.65          & 79.36         & \textbf{61.89} & 71.57          & 69.76 & 71.72 \\
ScaledSTE         & 9.66           & \textbf{72.53} & 63.99          & 79.57         & 61.32         & \textbf{72.40} & \textbf{69.96} & 72.12 \\
ScaleInit$_{0.1}$ & 28693.71       & 25.51          & 25.94          & 26.05         & 47.33          & 0.00           & 24.97 & 19.38 \\
LearnThresh       & 9.71           & 72.26          & \textbf{64.93} & \textbf{79.76}& 60.34         & 72.18          & 69.89 & \textbf{72.28} \\
\bottomrule
\end{tabular}}
\end{table}

The four non-divergent variants (NormScale / StdScale / ScaledSTE / LearnThresh)
fall within $0.07$\,PPL and $0.20$\,Avg-points of one another, so the design space is
robust to the choice of scaling rule, STE backward, and threshold. ScaleInit$_{0.1}$ catastrophically diverges
(PPL $\sim 28{,}000$) because $s_0 = 0.1$ violates Proposition~\ref{prop:soft-start}:
at step $0$ every adapted layer already carries a random full-rank perturbation
$\Delta W = 0.1\,\gamma\tilde{B}\tilde{A}$ of Frobenius norm $\approx 0.1\,\alpha\sqrt{d_{\text{out}}}$
($\approx 190$ on a Qwen2.5-7B query projection), applied before any
gradient signal, and training does not recover from it.
We adopt NormScale and LearnThresh as the two main-paper representatives.

\begin{table}[H]
\centering
\caption{Per-task evaluation time (seconds) for STE-variant ablation on Qwen2.5-7B.}
\label{tab:ste-time}
\small
\adjustbox{max width=\textwidth}{%
\begin{tabular}{lrrrrrrl}
\toprule
Method            & wikitext & mmlu  & arc\_c & hellaswag & tfqa & gsm8k & total \\
\midrule
QLoRA              & 252      & 667   & 186    & 1271      & 47   & 4874  & 2h 02m \\
NormScale          & 256      & 670   & 188    & 1288      & 48   & 5798  & 2h 17m \\
StdScale           & 255      & 669   & 187    & 1287      & 48   & 6150  & 2h 23m \\
ScaledSTE          & 256      & 671   & 186    & 1287      & 47   & 5847  & 2h 18m \\
ScaleInit$_{0.1}$  & 257      & 664   & 187    & 1279      & 47   & 7468  & 2h 45m \\
LearnThresh        & 256      & 671   & 189    & 1287      & 48   & 6143  & 2h 23m \\
\bottomrule
\end{tabular}}
\end{table}

\section{Scale Gate Dynamics Ablation}
\label{app:scale-dynamics}

The scale gate $s$ (initialized to 0) is central to BQ-LoRA's soft-start mechanism.
We investigate whether alternative dynamics for $s$---monotonic constraints, oscillating
modifiers, or alternating grow/decay schedules---can improve upon the default
free-gradient baseline. All experiments use Qwen2.5-7B-Instruct, 4-bit NF4, rank 16,
5{,}000 Alpaca samples, 1{,}000 steps, in the preliminary configuration ($\alpha = 1$ for
BQ-LoRA, warmup 100 steps). WT-PPL in this appendix comes from the preliminary in-house
evaluation (4-bit base: $10.37$), not from the lm-evaluation-harness word perplexity of the
main tables. Best is the lowest WT-PPL over the periodic evaluations during training, Alp
PPL is Alpaca validation perplexity capped at $50$, and Commit.\ is the fraction of latent
weights with $|w| > 0.5$.

\subsection{Monotonic decay}
$s_0 = 0.04$ (paper's converged value), apply
$s_{i+1} = \beta \min(s_{\text{prev}}, s_i)$ with $\beta < 1$.
All decay rates produce adapter death (Table~\ref{tab:mono-decay}).

\begin{table}[H]
\centering
\caption{Monotonic decay from $s_0=0.04$.}
\label{tab:mono-decay}
\small
\adjustbox{max width=\textwidth}{%
\begin{tabular}{lccccc}
\toprule
$\beta$ & WT-PPL & Best & Alp PPL & $s_{\text{final}}$ & Commit. \\
\midrule
0.999 & 10.37 & 10.22 & 50.0 & 0.014 & 0.016 \\
0.995 & 10.37 & 10.22 & 50.0 & 0.0003 & 0.009 \\
0.99  & 10.37 & 10.22 & 50.0 & ${\sim}0$ & 0.006 \\
\bottomrule
\end{tabular}}
\end{table}

\subsection{Monotonic growth}
$s_{i+1} = \beta \max(s_{\text{prev}}, s_i)$ with $\beta \geq 1$. $\beta = 1.0$ achieves
best PPL 10.04, $\beta = 1.005$ diverges (Table~\ref{tab:mono-grow}).

\begin{table}[H]
\centering
\caption{Monotonic growth from $s_0=0$.}
\label{tab:mono-grow}
\small
\adjustbox{max width=\textwidth}{%
\begin{tabular}{lccccc}
\toprule
$\beta$ & WT-PPL & Best & Alp PPL & $s_{\text{final}}$ & Commit. \\
\midrule
1.0 (max only) & 10.17 & 10.04 & 14.2 & 0.037 & 0.021 \\
1.001          & 10.22 & 10.05 &  8.4 & 0.072 & 0.019 \\
1.005          & 81{,}033 & 10.07 & 50.0 & 1.659 & 0.016 \\
\bottomrule
\end{tabular}}
\end{table}

\subsection{Oscillating scale modifier}
$s_{\text{eff}}(t) = s(t)(1 + A\,\mathrm{env}(t)\sin(2\pi t/P))$.
Damped envelope $1/(1+t/\tau)$ achieves PPL $9.95$, marginally above the free baseline
$9.94$ (Table~\ref{tab:oscillation}).

\begin{table}[H]
\centering
\caption{Oscillating scale modifier.}
\label{tab:oscillation}
\small
\adjustbox{max width=\textwidth}{%
\begin{tabular}{lccccc}
\toprule
Variant & WT-PPL & Best & Alp PPL & $s_{\text{final}}$ & Commit. \\
\midrule
Damp $A{=}0.5$, $P{=}200$ & 10.07 & \textbf{9.95} & 12.5 & ${\sim}0$ & 0.024 \\
Damp $A{=}0.5$, $P{=}300$ & 10.09 & 9.96 & 12.2 & ${\sim}0$ & 0.024 \\
Grow $A{=}0.3$, $P{=}200$ & 10.12 & 9.99 & 12.1 & ${\sim}0$ & 0.024 \\
\bottomrule
\end{tabular}}
\end{table}

\subsection{Alternating grow/decay (``breathing'')}
Alternate $s \leftarrow \beta_\uparrow \max(s_{\text{prev}}, s)$ and
$s \leftarrow \beta_\downarrow \min(s_{\text{prev}}, s)$. Net drift
$\beta_\uparrow \beta_\downarrow$ determines stability.

\begin{table}[H]
\centering
\caption{Alternating grow/decay with ratchet.}
\label{tab:breathing}
\small
\adjustbox{max width=\textwidth}{%
\begin{tabular}{lcccccc}
\toprule
$\beta_\uparrow / \beta_\downarrow$ & Net & WT-PPL & Best & Alp & $s_{\text{final}}$ \\
\midrule
1.1\;/\;0.9   & 0.99 & 10.33 & 10.14 & 50.0 & 0.0002 \\
1.25\;/\;0.8  & 1.00 & 10.10 & 9.96  & 11.0 & 0.004 \\
1.5\;/\;0.67  & 1.005 & 11.17 & \textbf{9.93} & 3.3 & 0.043 \\
2.0\;/\;0.5   & 1.00 & 10.08 & 9.95  & 9.6  & 0.016 \\
3.0\;/\;0.33  & 0.99 & 10.31 & 10.10 & 50.0 & 0.004 \\
2.0\;/\;0.8   & 1.60 & \multicolumn{4}{c}{(diverged, $s \to 10^{35}$)} \\
\bottomrule
\end{tabular}}
\end{table}

\subsection{Summary}
\begin{table}[H]
\centering
\caption{Best result per dynamics family.}
\label{tab:dynamics-summary}
\small
\adjustbox{max width=\textwidth}{%
\begin{tabular}{lccl}
\toprule
Family & Best PPL & $\Delta$ vs.\ free & Verdict \\
\midrule
Free (baseline)     & \textbf{9.94}  & ---    & \textbf{Optimal} \\
Oscillation (damped)& 9.95  & $+0.01$ & Comparable \\
Breathing (net$=1$) & 9.95  & $+0.01$ & Comparable \\
Mono grow ($\beta{=}1.0$)  & 10.04 & $+0.10$ & Slightly worse \\
Mono decay (all)    & 10.22 & $+0.28$ & Adapter death \\
\bottomrule
\end{tabular}}
\end{table}

Across $30+$ configurations spanning five families, no schedule ends training below the
free-gradient baseline. The $1.5/0.67$ breathing schedule reaches a lower minimum during
training ($9.93$) but ends at $11.17$. We therefore keep the zero-initialized, freely
learned gate.

\section{Rank and Step Ablation}
\label{app:rank-step}

\subsection{Rank ablation}
At rank 8 the BQ--QLoRA gap widens. At rank 32 QLoRA degrades (PPL $10.48 \to 11.89$)
while BQ remains stable. The $r{=}32$ NormScale GSM8K ($61.4$) is likely a single-seed
outlier; LearnThresh at $r{=}32$ achieves $73.2$.

\begin{table}[H]
\centering
\caption{Rank ablation on Qwen2.5-7B-Instruct.
Avg = mean of \{MMLU, ARC-C, Hella, TQA, GSM8K\}; Avg$^*$ excludes TQA.}
\label{tab:rank}
\small
\adjustbox{max width=\textwidth}{%
\begin{tabular}{llccccccccc}
\toprule
Rank & Method & PPL & MMLU & ARC-C & Hella & TQA & GSM8K & Avg & Avg$^*$ \\
\midrule
\multirow{3}{*}{r=8}
 & QLoRA       & 10.48 & 70.79 & 60.41 & 79.40 & 59.27 & 69.90 & 67.95 & 70.13 \\
 & NormScale   & \textbf{9.98}  & 72.35 & 64.51 & 79.86 & \textbf{61.63} & \textbf{74.98} & \textbf{70.67} & \textbf{72.93} \\
 & LearnThresh & 9.99  & \textbf{72.54} & \textbf{64.68} & \textbf{79.96} & 61.24 & 71.80 & 70.04 & 72.25 \\
\midrule
\multirow{3}{*}{r=32}
 & QLoRA       & 11.89 & 70.87 & 61.69 & \textbf{79.71} & 59.80 & 68.31 & 68.08 & 70.15 \\
 & NormScale   & \textbf{10.04} & \textbf{72.05} & 62.80 & 79.70 & 57.83 & 61.41 & 66.76 & 68.99 \\
 & LearnThresh & \textbf{10.04} & 71.68 & \textbf{63.23} & 79.59 & \textbf{61.03} & \textbf{73.24} & \textbf{69.75} & \textbf{71.94} \\
\bottomrule
\end{tabular}}
\end{table}

\subsection{Step ablation}
At $1{,}000$ steps QLoRA leads on PPL ($9.74$), MMLU and HellaSwag, while BQ already
leads on ARC-C, TQA-MC2, GSM8K and both averages. From $1{,}000$ to $3{,}000$ steps
(Table~\ref{tab:main-qwen}, 7B) QLoRA's PPL degrades from $9.74$ to $10.39$ and its MMLU
from $73.10$ to $70.47$, whereas BQ's PPL improves ($10.95 \to 9.87$ for NormScale,
$10.05 \to 9.86$ for LearnThresh): longer training overfits the FP16 adapter but not the
binary one.

\begin{table}[H]
\centering
\caption{Step ablation on Qwen2.5-7B-Instruct (1{,}000 steps, single seed).}
\label{tab:step}
\small
\adjustbox{max width=\textwidth}{%
\begin{tabular}{lcccccccc}
\toprule
Method & PPL & MMLU & ARC-C & Hella & TQA & GSM8K & Avg & Avg$^*$ \\
\midrule
QLoRA       & \textbf{9.74}  & \textbf{73.10} & 62.54 & \textbf{80.14} & 61.70 & 66.79 & 68.85 & 70.64 \\
NormScale   & 10.95          & 72.24          & 64.93 & 79.45          & \textbf{62.32} & 70.74 & 69.94 & 71.84 \\
LearnThresh & 10.05          & 72.35          & \textbf{65.19} & 79.64 & 62.06          & \textbf{71.19} & \textbf{70.09} & \textbf{72.09} \\
\bottomrule
\end{tabular}}
\end{table}

\section{Update geometry: Qwen2.5 vs.\ LLaMA-2}
\label{app:geometry}

To understand why BQ-LoRA leads on Qwen2.5 but not uniformly on LLaMA-2
(Table~\ref{tab:sota-llama}), we reconstruct the adapter updates $\Delta W$ of every adapted
projection from trained 7B checkpoints of both families (Alpaca, $3{,}000$ steps,
$\alpha = 32$; $112$ modules on Qwen2.5-7B, $128$ on LLaMA-2-7B) and measure three
quantities per module: the participation ratio, the top-1 energy share
$\sigma_{\max}^2/\|\Delta W\|_F^2$, and the relative update size
$\|\Delta W\|_F/\|W_{\text{base}}\|_F$.

\begin{table}[H]
\centering
\caption{Geometry of trained 7B adapter updates (mean over modules). The isotropic floor of
the top-1 share at $r = 16$ is $1/r = 0.0625$.}
\label{tab:geometry}
\small
\begin{tabular}{llcccc}
\toprule
Family & Method & PR (of 16) & Top-1 share & $\|\Delta W\|_F/\|W_{\text{base}}\|_F$ & mean $|s|$ \\
\midrule
\multirow{2}{*}{Qwen2.5-7B} & QLoRA     & 8.07  & 0.526 & 0.077 & -- \\
                            & NormScale & 15.92 & 0.079 & 0.162 & 0.0052 \\
\midrule
\multirow{2}{*}{LLaMA-2-7B} & QLoRA     & 8.36  & 0.505 & 0.064 & -- \\
                            & NormScale & 15.97 & 0.072 & 0.215 & 0.0086 \\
\bottomrule
\end{tabular}
\end{table}

\paragraph{The mechanism is family-general.}
QLoRA concentrates about half of the update energy in a single direction on both families
(top-1 share $0.51$--$0.53$), consistent with its participation ratio of about $8/16$.
BQ-LoRA is near-isotropic on both ($0.072$--$0.079$, within $15$--$26\%$ of the floor
$1/r$): binary signs under one scalar gate cannot allocate energy unevenly across
directions, which is the implicit regularizer of \S\ref{sec:theory}.

\paragraph{The demand on it is not.}
Under BQ-LoRA, where the direction is constrained and the gate is the only free magnitude,
LLaMA-2 learns $1.65\times$ larger gates and $1.33\times$ larger relative updates than
Qwen2.5. Under QLoRA, where directions are free, both families use small, concentrated
updates, slightly smaller on LLaMA-2 ($0.064$ vs.\ $0.077$), which rules out a naive
``LLaMA-2 needs larger updates'' explanation. Dynamic-rank training points the same way:
QDyLoRA's selected rank is $r^\ast = 4$ on all three seeds, while BQ-DyLoRA keeps
$r^\ast = 16$ on two of three (Table~\ref{tab:bqdylora}); concentrated FP16 updates
truncate cheaply, whereas near-isotropic binary updates spread energy over all $r$
directions. Our interpretation is that instruction-tuning a strongly pretrained base
(Qwen2.5, 18T tokens) is largely a re-weighting of existing behavior, for which bounded
isotropic updates are a good prior, whereas LLaMA-2 (2T tokens) needs more new behavior,
which concentrated FP16 directions (QLoRA, LoftQ) can inject more directly. This matches
the mixed LLaMA-2 aggregates.

\section{Adapter bit-width}
\label{app:bitwidth}

To test whether one bit is the right precision rather than a lucky point, we generalize the
sign quantizer to a $b$-bit symmetric uniform grid inside the identical framework, varying
only the quantizer. For $b = 1$ it is $\mathrm{sign}(\cdot)$; for $b > 1$, with
$\Delta = 2^{-(b-1)}$,
\begin{equation*}
q(x) = \mathrm{clamp}\bigl(\mathrm{round}(x/\Delta),\, -2^{b-1},\, 2^{b-1} - 1\bigr)\cdot\Delta,
\end{equation*}
so $b = 2$ gives $\{-1, -0.5, 0, +0.5\}$ and $b = 4$ gives 16 levels from $-1$ to $+0.875$.
The full $2^b$ codebook is used, so storage is exactly $b$ bits per parameter (a ternary
grid would carry $\log_2 3 = 1.58$ bits in 2 bits of storage). The backward pass is the
clipped STE, which at $b = 1$ is identical to NormScale's $|x| \leq 1$ mask; a self-test
verifies that the $b = 1$ path reproduces $\mathrm{sign}(\cdot)$, so the 1-bit row is
NormScale re-measured under the same protocol and seeds. Latent initialization, the soft
start, rank, $\alpha$, data, steps and evaluation are held fixed (Qwen2.5-7B, Alpaca,
3 seeds).

\begin{table}[H]
\centering
\caption{Adapter bit-width on Qwen2.5-7B (Alpaca, 3 seeds, mean$\pm$std). Storage in KB;
Avg = mean of the four task metrics.}
\label{tab:bitwidth}
\small
\adjustbox{max width=\textwidth}{%
\begin{tabular}{lrrcccccc}
\toprule
Precision & KB & bits/param & PPL$\downarrow$ & MMLU & ARC-C & Hella & TQA & Avg \\
\midrule
1-bit (NormScale) & \textbf{1303} & 1.03 & 9.91\,$\pm$\,0.22 & 72.35\,$\pm$\,0.10 & \textbf{63.94\,$\pm$\,0.36} & 79.65\,$\pm$\,0.24 & \textbf{60.97\,$\pm$\,0.54} & \textbf{69.23} \\
2-bit             & 2388 & 1.94 & 9.92\,$\pm$\,0.05 & \textbf{72.75\,$\pm$\,0.22} & 62.63\,$\pm$\,0.09 & 79.96\,$\pm$\,0.34 & 59.96\,$\pm$\,1.20 & 68.83 \\
4-bit             & 5150 & 4.18 & \textbf{9.83\,$\pm$\,0.02} & 72.73\,$\pm$\,0.03 & 62.97\,$\pm$\,0.76 & \textbf{80.10\,$\pm$\,0.11} & 59.05\,$\pm$\,0.72 & 68.71 \\
\bottomrule
\end{tabular}}
\end{table}

Quality is not monotone in bit-width. One bit leads on ARC-C ($+1.0$--$1.3$) and TQA
($+1.0$--$1.9$), ties 2-bit on PPL (4-bit's $-0.08$ lies within 1-bit's seed spread), and
trails only on MMLU ($-0.4$) and HellaSwag ($-0.3$ to $-0.45$), giving the highest average
of the three at $1.8$--$4\times$ less storage. Training cost does not depend on $b$, since
the latents are FP32 for every bit-width.

\section{Training cost}
\label{app:train-cost}

We profile training under identical conditions: Qwen2.5-7B NF4 base, $r = 16$ on
$\{q,k,v,o\}$, batch $4 \times$ sequence length $1024$, AdamW, gradient checkpointing, a
single A100-80GB, and 50 measured steps after 10 warm-up steps.

\begin{table}[H]
\centering
\caption{Training cost vs.\ QLoRA on Qwen2.5-7B.}
\label{tab:train-cost}
\small
\adjustbox{max width=\textwidth}{%
\begin{tabular}{lccccc}
\toprule
Method & Peak mem (GB) & Step time (ms) & Min / 1k steps & Trainable params & Optimizer state \\
\midrule
QLoRA       & 14.26 & 1079          & 18.0 & 10.09\,M              & 77\,MB \\
NormScale   & 14.26 & 1108 (+2.7\%) & 18.5 & 10.09\,M (+112 scales)  & 77\,MB \\
LearnThresh & 14.26 & 1121 (+3.8\%) & 18.7 & 10.09\,M (+336 scalars) & 77\,MB \\
\bottomrule
\end{tabular}}
\end{table}

BQ-LoRA's FP32 latents occupy the same memory as QLoRA's adapter weights, the optimizer
state is identical, and peak memory is dominated by components shared across methods (the
quantized base, activations and logits). The only measurable overhead is $2.7$--$3.8\%$
step time from the sign and STE operations. The $13$--$16\times$ reduction therefore
applies to inference-time adapter storage and serving, not to training.

\section{Implementation Details}
\label{app:impl}

\paragraph{Hardware.} A100-80GB (Ablations), H200-141GB (dataset ablation, 1.5B/7B/70B/72B).

\paragraph{Hyperparameters.}
LR $2 \times 10^{-4}$ with linear warmup over 300 steps, AdamW, weight decay $0.01$,
$\alpha = 32$ for every method in the final runs (QLoRA scaling $\alpha/r$, BQ-LoRA
scaling $\gamma = \alpha/\sqrt{r d_{\text{in}}}$; only the preliminary training-dynamics runs
of Appendix~\ref{app:scale-dynamics} used $\alpha = 1$ for BQ-LoRA), global gradient-norm clipping $1.0$, batch size $4 \times $ accum $4$ ($8 \times 2$ for the
BQ-DyLoRA runs, the same effective batch), max sequence length $128$, 3{,}000 steps,
dropout $0.05$. Latent weights initialized $\mathcal{U}(-0.5, 0.5)$; FP32 adapter forward.

\paragraph{Common pitfalls (and fixes).}
\begin{enumerate}
    \item Custom perplexity loops over short non-overlapping chunks inflate WikiText-2
    perplexity by $\sim 3$ points; we report the lm-evaluation-harness \texttt{wikitext}
    word perplexity.
    \item GSM8K uses the stock lm-evaluation-harness \texttt{gsm8k} task (8-shot, full
    chain-of-thought exemplars, strict-match). Custom formats that keep only the final
    number in the exemplars or cap generation at 64 tokens suppress chain-of-thought and
    are not comparable.
    \item For Qwen and LLaMA tokenizers, \texttt{pad\_token = eos\_token}; padding must
    be masked to $-100$ in labels or the model learns to predict EOS at every padding
    position.
    \item Leaving \texttt{use\_cache=False} on during \texttt{model.generate()} causes
    $O(N^2)$ KV recomputation.
    \item BitsAndBytes-quantized layers do not expose \texttt{base\_layer.weight.shape};
    use \texttt{out\_features} / \texttt{in\_features}.
    \item When combining LoRA with gradient checkpointing on the base model, call
    \texttt{model.enable\_input\_require\_grads()}.
\end{enumerate}

\end{document}